# FibreCastML: An Open Web Platform for Predicting Electrospun Nanofibre Diameter Distributions for Biomedical Applications

**Elisa Roldán[1]*, Kirstie Andrews[1], Stephen M. Richardson[2], Reyhaneh Fatahian[1], Glen Cooper[3], Rasool Erfani[1], Tasneem Sabir[4], Neil D. Reeves[5]**

[1] Department of Engineering, Faculty of Science & Engineering, Manchester Metropolitan University, Manchester M1 5GD, UK

[2] Division of Cell Matrix Biology and Regenerative Medicine, School of Biological Sciences, Faculty of Biology, Medicine and Health, The University of Manchester, Stopford Building, Oxford Rd, Manchester, M13 9PT

[3] School of Engineering, University of Manchester, Manchester, M13 9PL, UK

[4] Manchester Fashion Institute, Faculty of Arts & Humanities, Manchester Metropolitan University, Manchester M15 6BG, UK

[5] Lancaster Medical School, Faculty of Health and Medicine, Lancaster University, Lancaster LA1 4YW, UK

*** Correspondence:**
Elisa.Roldan-Ciudad@mmu.ac.uk

## Abstract

INTRODUCTION: Electrospinning is a scalable technique for generating fibrous scaffolds with tunable micro- and nanoscale architectures for tissue engineering, drug delivery, and wound care. Machine learning (ML) has emerged as a powerful tool to accelerate process optimisation; however, existing models typically predict only mean fibre diameters, overlooking the entire diameter distribution that governs scaffold functionality and biomimicry. This study introduces FibreCastML, the first open-access, distribution-aware ML framework that predicts full fibre diameter spectra from routinely reported processing parameters and provides interpretable insights into parameter influence.

METHODS: A comprehensive meta-dataset of 68,538 fibre-diameter measurements from 1,778 studies across 16 biomedical polymers was curated. Six standard input parameters (solution concentration, voltage, flow rate, tip-to-collector distance, needle diameter, and rotation speed) were used to train seven ML learners (linear model, elastic net, decision tree, multivariate adaptive regression splines, k-Nearest Neighbours, random forest, and radial-basis Support Vector Machine) under nested cross-validation with leave-one-study-out external folds to ensure generalisable performance. Model interpretability combined variable importance, SHapley Additive exPlanations (SHAP), correlation matrices, and 3D parameter maps. The FibreCastML web app integrates these capabilities with out-of-range detection, solvent suggestions, and automated Excel reports.

RESULTS: Non-linear and local learners consistently outperformed linear baselines, achieving $R^2 >$ 0.91 for polymers such as cellulose acetate, Nylon-6, Polyacrylonitrile, polyD,L-lactide, Polymethyl methacrylate, Polystyrene, Polyurethane, Polyvinyl alcohol (PVA), and Polyvinylidene fluoride.

Concentration emerged as the most influential variable globally. The FibreCastML app returns polymer-specific distribution plots, predicted-vs-observed diagnostics, feature importance and correlations, and transparent metrics ($R^2$, RMSE, MAE) for user-defined settings. In an experimental case study using different electrospinners and microscopies, predicted diameter distributions closely matched experimental measurements (Kolmogorov–Smirnov $p > 0.13$ and overlap coefficient of 84%).

DISCUSSION: By shifting from mean-centric to distribution-aware modelling, this work establishes a new paradigm for electrospinning design. FibreCastML enables reproducible, sustainable, and data-driven optimisation of scaffold architecture, bridging experimental and computational domains. Openly available, it empowers laboratories worldwide to perform faster, greener, and more reproducible electrospinning research, advancing sustainable nanomanufacturing and biomedical innovation.

# 1 Introduction

Electrospinning is a versatile electrohydrodynamic process that enables the fabrication of micro- and nano- polymer fibres with architectures that emulate key features of the extracellular matrix, underpinning applications in biomedical engineering such as regenerative medicine and tissue engineering [1]; wound care and infection control [2,3]; drug, protein, gene and vaccine delivery [4,5]; medical device integration [6]; diagnostics, biosensing and wearables [7,8]; viral filtration [9]; or research tools such as organ-on-chip [10]. The process initiates when the applied electric field overcomes surface tension at the pendant droplet, forming the characteristic Taylor cone; once a critical field is reached, a charged jet is ejected, stretches, and undergoes bending/whipping instabilities before solidifying on the collector. The onset and stability of this cone, jet regime, and thus the fibre population that ultimately forms, depend on a coupled set of solution (including concentration, viscosity, molecular weight, surface tension, or conductivity), process (voltage, flow rate, needle diameter, type of collector, revolutions of the mandrel, or tip-to-collector distance), and ambient parameters (such as relative humidity or temperature) [11]. Together, these factors control morphology (fibre-diameter distribution, inter-fibre separation), topography (alignment, surface roughness), and emergent mechanical properties (Young's modulus, ultimate tensile strength or strain at break) of the electrospun scaffolds [12]. However, establishing the ideal combination of those parameters, to obtain the desired properties, is one of the biggest challenges in electrospinning [13].

One of the properties most studied in electrospinning is the diameter of the fibre, and how the process parameters affect to it. While average fibre diameter is an important design lever, affecting surface area, porosity, mechanics, and cell–material interactions, there is mounting evidence that diameter distributions govern mesoscale behaviour, including local stiffness heterogeneity, transport percolation, and fibre–fibre contact statistics [14–16]. This distinction is decisive for biomimicry, where healthy ligaments exhibit a robust bimodal collagen-diameter spectrum with two well-defined peaks, whereas injured ligaments collapse toward a unimodal distribution, with measurable consequences for function and mechanobiology [17,18]. Designing electrospun scaffolds to reproduce such spectra therefore requires predictive tools that map processing choices to the full conditional distribution of fibre diameters, not merely to a single expected mean value.

Conventional optimisation, one-factor-at-a-time experiments and classical designs of experiments (e.g., response-surface, Box–Behnken, central composite) improves efficiency but remains costly, time-consuming, narrow in scope, and largely mean diameter-centred [19]. Recent machine-learning (ML) approaches have achieved accurate point predictions of diameter from solution chemistry, processing, and equipment variables [20–22]. However, most published models return a single number, which is insufficient when the design target is a specified shape of the diameter spectrum (e.g., unimodal vs bimodal) that must be propagated into microstructure-aware simulations of tissue behaviour.

This article advances electrospinning from mean-centred optimisation to distribution-aware design, presenting the first framework to predict the entire distribution of electrospun fibre diameters conditioned on polymer–solvent composition and process parameters, explicitly capturing unimodal and bimodal behaviours essential for mimicking healthy and injured tissue microstructures.

Central to this contribution is a novel, comprehensive database curated from 1,778 independent studies, spanning 16 polymers and comprising 68,538 fibre-diameter observations. The database covers the following polymers: Cellulose acetate (CA), Gelatin, Polyamide 6 (Nylon 6), Polyacrylonitrile (PAN), Polycaprolactone (PCL), polyD,L-lactide (PDLLA), Polyether ether ketone (PEEK), Polyethylene terephthalate (PET), Polylactic acid (PLA), Polymethyl methacrylate (PMMA), Polystyrene (PS), Polyurethane (PU), Polyvinyl alcohol (PVA), Polyvinylidene fluoride (PVDF), Polyvinylpyrrolidone (PVP) and poly γ-glutamic acid (γ-PGA). In electrospun form, these materials have been deployed for biomedical purposes including antibacterial wound dressings [23–30], tissue engineering [31–37] and regenerative medicine [38–40]. Recently, case studies of these nanostructures have been highlighted as key metamaterials and metasurfaces for health and wellbeing, therapeutics and sustainability in a policy report [41]. Collectively, these exemplars illustrate that each material has a credible electrospun route into biomedical applications.

A second key contribution is translational: FibreCastML is the first open-access web application that allows users to specify electrospinning parameters and obtain polymer-specific predictions of the full fibre-diameter distribution rather than a single average. The application (i) flags inputs outside the observed domain of the database, (ii) visualises predicted vs observed behaviour and metrics performance of seven supervised ML algorithms for the 16 polymers, (iii) complements model benchmarking with interpretable analyses (variable importance and SHAP) and parameter correlations, (iv) offers solvent-system suggestions by proximity in parameter space, and (v) provide all the results in a Excel worksheet.

The impact of this work is a step-change in how electrospinning is done: instead of iterating blindly on mean diameter, researchers can plan and justify experiments up front against the full fibre-diameter spectrum that governs function. Concretely, the framework (i) reduces trial-and-error, solvent consumption, and material waste by letting teams screen conditions digitally; (ii) improves reproducibility and auditability through standardised diagnostics, out-of-range warnings, and an exportable Excel report; (iii) raises efficiency by pointing to the most controllable levers (e.g., concentration) and by proposing historically successful solvent systems obtained from a large-scale database of 68,538 observations; and (iv) democratises access with an open app so resource-constrained groups can reach biomimetic targets without costly optimisation campaigns. In addition, to illustrate practical usability, an experimental case study was performed. The result is a safer, more sustainable, and faster translation of electrospun scaffolds for tissue engineering, wound care, and drug delivery, backed by a large, curated evidence base and rigorous validation.

# 2 Methods

## 2.1 Data collection

A comprehensive dataset was created by reviewing research literature in the Scopus and Google Scholar databases. The search was conducted using the keywords “electrospinning” & popular polymers used in electrospinning for biomedical applications such as Cellulose acetate (CA), Gelatin, Polyamide 6 (Nylon 6), Polyacrylonitrile (PAN), Polycaprolactone (PCL), polyD,L-lactide (PDLLA), Polyether ether ketone (PEEK), Polyethylene terephthalate (PET), Polylactic acid (PLA), Polymethyl methacrylate (PMMA), Polystyrene (PS), Polyurethane (PU), Polyvinyl alcohol (PVA), Polyvinylidene fluoride (PVDF), Polyvinylpyrrolidone (PVP) and poly γ-glutamic acid (γ-PGA). Only full reproducible articles with single polymers and reporting fibre distributions in their experimental designs were included in the dataset. In addition to this search, the dataset was completed with two published datasets [42,43] which, in total, yielded 1,778 studies.

The dataset consist of the following input fields: The document identifier (DOI), polymer, solvent1, solvent2, solvent3, solvent1_ratio (%), solvent2_ratio (%), solvent3_ratio (%), concentration (%), needle diameter, type of collector, rotation speed (rpm), voltage (kV), flow rate (ml/h), tip-to-collector distance (cm), temperature (°C) and humidity (%). DOI was retained solely for provenance tracking and auditability and was not used as a predictive feature during model training or inference. The output variable is the distribution of the diameter of the fibres obtained from the 1,778 studies, comprising a total of 68,538 fibre-diameter observations.

## 2.2 Data preprocessing

Data were imported with all fields initially read as text to prevent implicit type coercion. Numeric fields were parsed using a custom routine that removed non-numeric characters, reconciled European/US decimal conventions (comma vs dot), and converted to numeric; non-parsable entries became missing. The entire dataset was checked for missing values, outliers, and inconsistencies. Observations with non-finite fibre diameter were discarded and rows with missing values were dropped (no imputation).

Preprocessing and normalisation were implemented using the caret library’s recipes pipeline. For each polymer-specific dataset, the recipe included (i) removal of zero-variance predictors (step_zv on all predictors) and (ii) z-normalisation (step_normalize) of all numeric predictors to zero mean and unit variance, while leaving the outcome (fibre diameter, in nm) on its original scale. To prevent information leakage, these steps were estimated independently within each resampling split: the centring and scaling parameters were computed using only the training folds in the repeated 5-fold cross-validation and then applied to the corresponding validation folds and to any new prediction inputs, following previous studies [44,45](Roldán et al., 2023b).

All preprocessing, ML modeling, and reported library functions were implemented in R 4.3.0 using RStudio 2023.03.1.

## 2.4 Prediction models

The supervised learning task is defined on a polymer-specific subset of the data in which six process variables (solution concentration, needle diameter (g), rotation speed, voltage (kV), flow rate (ml $h^{-1}$), and tip-to-collector distance (cm)) serve as inputs, and the fibre diameters are the output. Models are trained with standardised predictors (zero-variance removal and z-score normalisation) using the recipes and caret ecosystem. Seven complementary learners were evaluated to span simple, interpretable models through flexible nonlinear methods. Ordinary least squares (lm) were used as a transparent linear baseline, relating each predictor to the outcome with interpretable coefficients. Elastic net (glmnet) extended this baseline by adding L1/L2 regularisation to stabilise estimates under multicollinearity and shrink or discard weak signals [46]. Decision trees (rpart) partitioned the data into if–then rules, providing human-readable logic but with a tendency to overfit [47]. Random forests (ranger) mitigated that risk by averaging many trees grown on resampled data and random subsets of variables, improving accuracy and robustness [48]. A radial-basis support vector machine (kernlab's svmRadial) modelled smooth nonlinear relationships by mapping inputs to a higher-dimensional space and maximising the margin [49]. K-nearest neighbours (knn) produced predictions based on the most similar cases in the training set, offering a simple, assumption-light local method [50]. Finally, multivariate adaptive regression splines (earth/MARS) captured curved effects by stitching together piecewise-linear segments with automatically placed knots, maintaining a degree of interpretability [51].

Hyperparameters were tuned on compact, deterministic grids appropriate to heterogeneous experimental datasets: elastic net over $\alpha \in \{0,0.25,0.5,0.75,1\}$ with $\lambda$ on a logarithmic grid; random forest with mtry spanning 1…p, min.node.size$\in\{1,5,10\}$ and 500 trees; SVM-R with sigma initialised by sigest and $C\in\{0.25,0.5,1,2,4\}$; decision tree with cp on 10^{−4}…10^{−1}; kNN with $k\in\{3,5,7,9,11\}$; MARS with degree$\in\{1,2\}$ and nprune$\in\{5,10,15,20,25\}$. All training used fixed random seeds to ensure exact reproducibility, and plotting and reporting were handled with ggplot2, dplyr/tidyr, readxl, and openxlsx.

Model selection and performance estimation followed a nested cross-validation design to avoid optimistic bias following previous studies [44,45]. An inner loop of five-fold cross-validation with two repeats was used to tune each model's hyperparameters (via caret::trainControl), a choice that balances bias and variance of the tuning criterion while remaining computationally tractable in an interactive setting; repeating the folds stabilises the selected configuration in small-n regimes typical of polymer-specific datasets. The outer loop adopted a leave-one-study-out K-fold scheme, where K is the number of independent studies available for the polymer (Figure 1). This grouping (implemented via study-level folds) yields an honest assessment of generalisation to unseen studies, minimises leakage arising from repeated or near duplicate conditions within a study, and mirrors the app's real-world use in which users often extrapolate to new experimental batches. It is also worth noting that, as all modelling was conducted on polymer-specific subsets, each polymer is evaluated only within its own distribution, ensuring that no imbalance data can affect training, hyperparameter selection, or interpretability. This also eliminates the need for categorical encoding: the polymer variable is not included as a predictor and each model learns only from homogeneous material behaviour.

Aggregating these out-of-fold predictions provides the empirical distribution of modelled fibre diameters used in the interface's histograms and predicted-versus-observed plots, delivering both a calibrated point prediction and an interpretable sense of spread for decision-making.

Performance metrics were computed on the out-of-fold predictions. Coefficient of determination ($R^2$), square root of the mean squared error (RMSE) and mean absolute error (MAE) were calculated following the equations (1-3):

$$R^2 = 1 - \frac{\sum_{i=1}^{n}(y_i - \hat{y}_i)^2}{\sum_{i=1}^{n}(y_i - \bar{y})^2} \quad (1)$$

$$MAE = \frac{1}{n}\sum_{i=1}^{n}|y_i - \hat{y}_i| \quad (2)$$

$$RMSE = \sqrt{\frac{1}{n}\sum_{i=1}^{n}(y_i - \hat{y}_i)^2} \quad (3)$$

Where $y_i$ is the actual value, $\hat{y}_i$ is the predicted value, $\bar{y}$ is the mean of the actual values, and n is the number of samples.

To quantify model stability, all performance metrics ($R^2$, RMSE, MAE) are reported as mean ± standard deviation across outer cross-validation folds, which is widely used in machine-learning evaluation to characterise fold-to-fold variability.

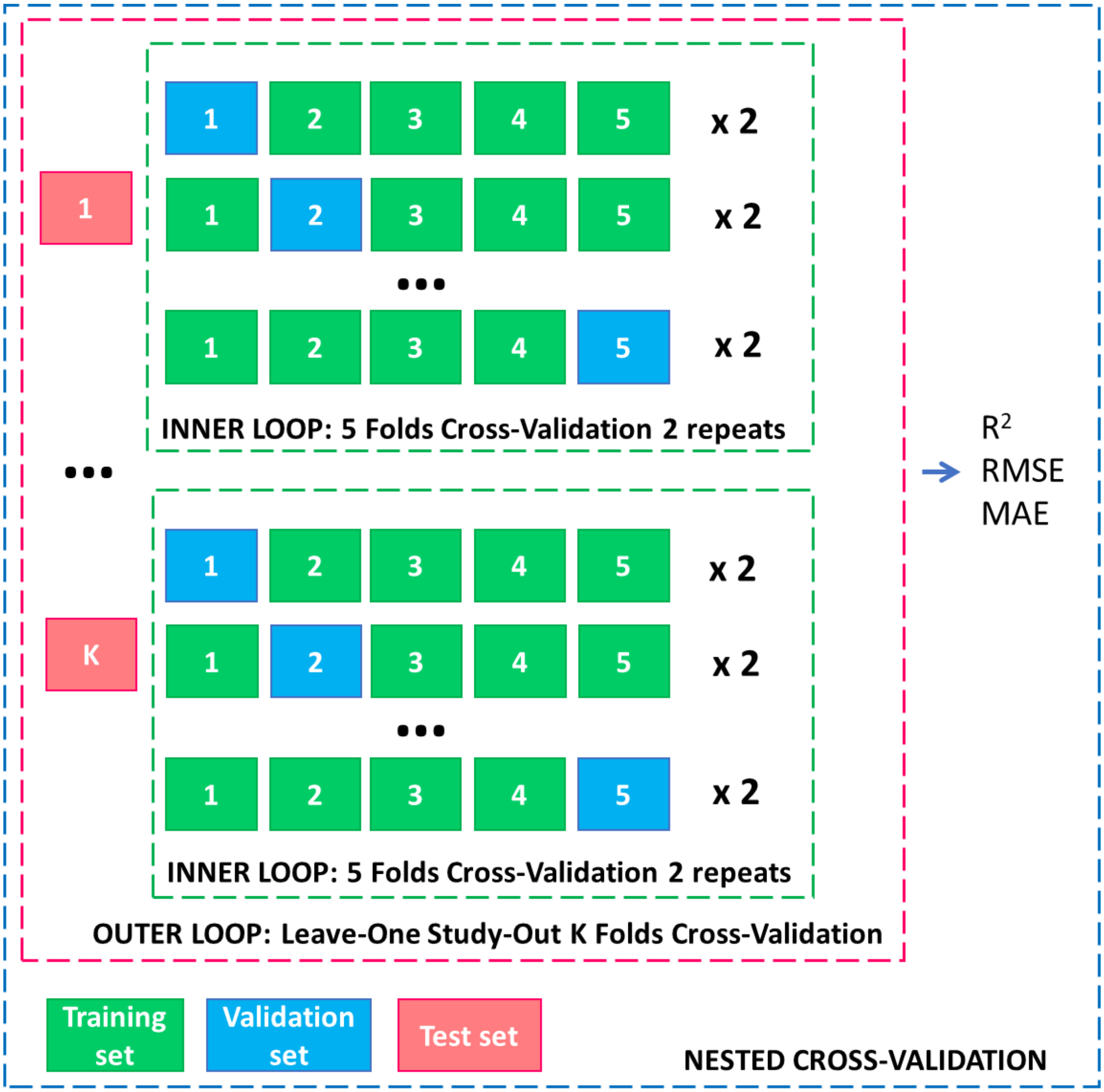

**Figure 1**. Validation of the models.

### 2.5 Importance of the features on the diameter of the fibres

To clarify which process parameters most strongly shape the predicted fibre-diameter distribution, two complementary quantities are reported: model-specific variable importance and model-agnostic SHAP (SHapley Additive exPlanations) values. Variable importance provides a global ranking of predictors, across all observations, according to how much each contributes to the model's predictive capability under its own learning mechanism. In practical terms, higher-importance variables are those to which the fitted model is most sensitive when explaining variability in fibre diameter; hence, they are prime candidates for process control and experimental refinement. SHAP values, in contrast, are local attributions defined for each individual prediction. They quantify how much each feature pushes a prediction above or below a baseline expectation (the model's average output on a reference background). Positive SHAP values indicate that a feature configuration increases the predicted diameter; negative values indicate a decrease. Aggregated across specimens, the spread and central tendency of SHAP values illuminate not only which variables matter but how they matter (directionality and effect size), offering an interpretable link between parameter settings and the modelled distribution of diameters.

In the FibreCastML application, variable importance is computed with the caret framework's native varImp routine for the selected algorithm, yielding a single overall score per feature and visualised as a ranked bar chart (top 20). SHAP values are estimated with fastshap, which approximates Shapley attributions for the trained model by repeatedly perturbing features on a background sample (up to 200 rows drawn from the polymer-specific training set; 50 Monte Carlo simulations per feature) and passing them through a prediction wrapper identical to the one used for inference. SHAP results are summarised by ordering features by mean absolute SHAP magnitude and displaying a jittered dot plot for the six most influential variables, thereby showing both the central effect and its variability across cases. Together, the global ranking (importance) and the local attributions (SHAP) provide a robust, multi-scale explanation of the factors governing the distribution of fibre diameters in silico. It is noted that importance metrics and SHAP interpretations can be attenuated by multicollinearity and by model choice; accordingly, 3D response-surface plots and a correlation heat map is presented alongside these diagnostics to contextualise overlaps among predictors and to support cautious, experiment-driven conclusions.

## 2.6 FibreCastML Open Web Platform

Shiny is an R framework for building interactive web applications directly from R scripts. It uses a reactive programming model that links a declarative user interface to a server that performs statistical computation. Inputs such as selectors, numeric fields, and buttons automatically update outputs (tables, figures, and text) whenever data or parameters change. This approach supports reproducibility and auditability of scientific analyses and enables seamless local or web deployment for collaborative use.

FibreCastML is a Shiny interface that predicts electrospun fibre diameters for a user-selected polymer and reports model performance, interpretability outputs, data diagnostics, and a reproducible summary. On initialisation, the application loads a comprehensive dataset of 68,538 fibre-diameter observations created by this research group and located alongside the app, and performs the preprocessing detailed in section 2.2. The app then detects which modelling engines are available on the host and offers only those that can run there (e.g., linear model, elastic net, random forest, radial-basis SVM, decision tree, k-nearest neighbours, and MARS), with a sidebar note indicating what is available on that server.

The left-hand panel provides the control surface. The user selects a "Polymer" from those detected in the dataset and a "Collector type". They then enter the operative electrospinning parameters as numbers: solution concentration, needle diameter (g), rotation speed, voltage (kV), flow rate (ml/h), and tip-to-collector distance (cm). A drop-down allows the choice of learning algorithm. A brief instruction reminds the user to complete all numeric fields before pressing "Run prediction". During computation, the interface displays "WAIT… PROCESSING"; upon completion, it switches to "RESULTS IN PREDICTION & METRICS TAB". A download control is always visible to export a comprehensive Excel report after a run has finished.

Outputs are organised into two tabs. The "STATUS" tab provides immediate, decision-support feedback prior to reviewing the full results. It first echoes the current application state. It then offers a polymer-specific solvent recommendation derived from historical records: within the subset matching the selected polymer, the app identifies candidate experiments that are close to the user's settings after scaling each parameter by its observed variability. Candidates are scored using a convex combination of parameter proximity and closeness of fibre diameter to the current prediction, and the most frequent solvent triplet is summarised. Where ratio columns exist, median percentages for each solvent among the top candidates are also returned, proposing both composition and approximate proportions. Finally, the tab lists any user-entered parameters that fall outside the observed range for that polymer and displays the corresponding minima and maxima; if all values are within range, this is stated explicitly.

The "PREDICTION & METRICS" tab delivers the analytical core. When "Run prediction" is clicked, the server filters the dataset to the chosen polymer and constructs a modelling table with the six process parameters as predictors and the fibre diameter as the target. Once fitted, the app computes and displays a single prediction of fibre diameter for the user's settings. It then shows a univariate distribution of out-of-sample predicted diameters arising from cross-validated fits and overlays a dotted vertical line at the user's predicted value with a textual annotation, situating the configuration within the model's typical prediction range. A concise performance table reports RMSE, mean absolute error (MAE), and the cross-validated coefficient of determination ($R^2$) from the cross-validation. Where the learner provides interpretable parameters, most directly in ordinary least squares and elastic net, the app prints a coefficient table with estimated effects; when a model lacks transparent coefficients, the interface states this clearly.

Diagnostics and interpretability are extended through additional figures. A "Predicted vs Observed" scatter plots held-out predictions against actual fibre diameters with a unity line to reveal systematic bias or dispersion. Variable importance is computed via the modelling framework's native mechanism and rendered as a horizontal bar chart for up to the top twenty predictors, enabling rapid identification of the most influential drivers. The app also estimates SHAP values using a background sample of the feature space and presents a dot-plot summary ordered by mean absolute SHAP magnitude for the six most influential variables, providing model-agnostic explanations that connect local and global behaviour. Finally, a correlation heat map spanning all predictors and the target is displayed with numeric overlays, aiding the detection of collinearity and linear associations in the polymer-specific training subset.

All outputs are reproducibly exportable via a structured Excel workbook generated on demand. The report contains a "Summary" sheet capturing the user's inputs, the chosen model, and the predicted diameter, as well as the solvent recommendation; an "Out_of_Range" sheet enumerating any parameter excursions relative to the empirical domain for the polymer (note that although the tool flags certain user-defined conditions as 'out of range,' this message simply indicates that the parameters fall outside the values commonly reported in the literature for stable Taylor cone formation, and the model still performs predictions and generates all interpretability outputs); data-rich sheets for cross-validated predictions, prediction distribution, metrics, coefficients (when available), variable importance, SHAP summaries (when available), and the correlation matrix; and embedded, publication-ready versions of the key figures (prediction distribution, predicted-versus-observed scatter, variable importance chart, SHAP summary, and correlation heat map).

The followed procedure for the whole study can be found in Figure 2.

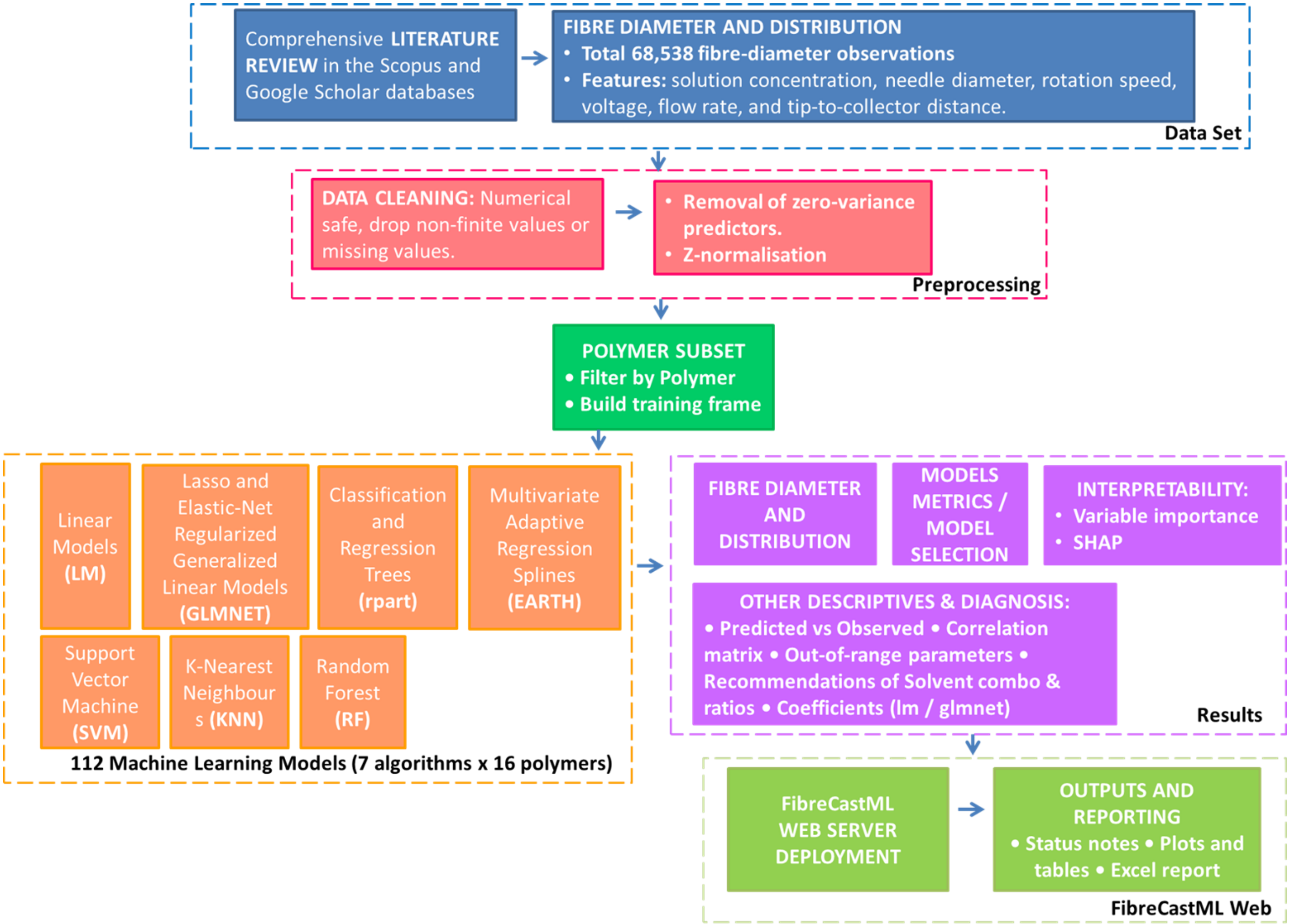

**Figure 2**. Outline of the followed methodology from dataset creation to open web platform.

## 2.7 Experimental case study

Following rigorous nested cross-validation, which decoupled hyperparameter optimisation from performance evaluation to ensure unbiased estimation of model generalisation and minimise optimistic bias, an experimental case study was conducted to demonstrate the practical applicability of the FibreCastML platform. This case study was not intended as an additional validation step, but rather as a proof-of-use illustrating real-world deployment of the trained models. For the experimental case study, an external dataset was generated. All new data were entirely independent from the training database; no images, or samples used for model development were included in this validation study. To ensure the robustness of the case study, two different electrospinning devices, and two separate scanning electron microscopes (SEMs) were employed.

Electrospun scaffolds were produced on different days using two different electrospinning systems:

(i) A TL-01, NaBond (China),
(ii) a Spraybase® (Ireland), and

PVA with average molecular weight 30,000-70,000 (commonly used in electrospinning) were included in this experimental case (#P8136, Sigma Aldrich, UK). All PVA-based solutions were produced by dissolving the polymer in distilled water ($dH_2O$) at 12% w/w by heating to 80 °C with stirring until a homogeneous solution formed. A 10 ml syringe was loaded with 5 ml of polymer solution, and it pumped with at 1 ml/h flow rate through an 18 G needle. A voltage of 20 kV was applied between the needle tip and the collector to generate the electrostatic field. Fibres were deposited onto aluminium foil affixed to a rotating drum collector operating at 2000 rpm. The tip-to–collector distance was set to 10 cm. Electrospinning was conducted at room temperature (25 °C).

Samples were sputter-coated with Au/Pd using a SC7640 sputter coater (Quorum Technologies Ltd., Kent, UK) prior to visualisation. Coating was performed at 20 mA and 0.8 kV for 120 s, yielding a nominal thickness of 32.6 nm according to the instrument specification. Two different SEMs were used for image acquisition: a Benchtop SEM (Hitachi TM4000Plus, Hitachi High-Tech Europe GmbH) and field emission scanning electron microscope (Zeiss Supra 40, FE-SEM, Carl Zeiss SMT Ltd., Cambridge, UK) to ensure that the model remained robust across different SEM devices. SEM micrographs were acquired at an accelerating voltage of 2 kV, a working distance of approximately 6 mm, and magnifications up to ×30,000.

Fibre diameter (Ø) was quantified in AxioVision SE64 Rel. 4.9.1 (Carl Zeiss SMT Ltd., Cambridge, UK) following a previous study [52]. For each sample, twenty fibres were measured from representative high-magnification fields. To guaranty reproducibility across electrospinner systems, three replicates were performed on different days and two samples per replicate were evaluated.

Statistical analyses were performed to assess the usability of FibreCastML. All statistical analyses were performed to compare the FibreCastML-predicted fibre diameters with the experimentally measured diameters obtained from the TL-01 and Spraybase® electrospinning systems. All analyses were conducted using a significance threshold of $p < 0.05$.

Prior to inferential testing, data distributions for each group were assessed for normality using the Shapiro–Wilk test, which indicated significant deviations from normality; therefore, non-parametric methods were selected. Descriptive statistics (median, standard deviation, interquartile range, minimum and maximum values) were calculated for each dataset to characterise central tendency and dispersion.

In addition to providing a single point prediction for the fibre diameter, the model explicitly quantifies predictive uncertainty through a residual-bootstrap Monte Carlo procedure tailored to the statistical behaviour of the experimental dataset. In case of PVA, the dataset contains 6,610 PVA records spanning a wide range of electrospinning configurations; thus, the dispersion observed in the data reflects both the intrinsic variability of PVA and the system's sensitivity to diverse processing conditions. Consequently, each prediction should be interpreted not as a single deterministic value, but as a draw from the conditional distribution of fibre diameters for the hypothetical population defined by the user-selected inputs. To approximate this conditional predictive distribution, the model first computes the point prediction $\hat{y}$ for a new set of conditions and then generates 100 Monte Carlo realisations of the predicted diameter using a residual bootstrap of the cross-validated errors following equation 4,

$$\tilde{y}_i = \hat{y} + \varepsilon_i^* \qquad (4)$$

where $\varepsilon_i^*$ are sampled with replacement from the empirical residual distribution. This non-parametric procedure yields an empirical predictive distribution conditioned on the selected inputs, informed by the variability observed across the 6,610 experimental PVA configurations. The resulting ensemble is then used for statistical comparison with real scaffold measurements (Kolmogorov Smirnov, U Mann–Whitney, t-test included solely as a complementary assessment of mean differences, overlap coefficient, Kullback–Leibler divergence, Wasserstein distance) and for visual distributional diagnostics. Integrating residual-bootstrap uncertainty quantification directly into an interactive, process-aware electrospinning prediction tool is, to our knowledge, novel in this field and moves beyond standard practice based solely on point estimates or global error metrics, enabling a more transparent, data-driven and risk-aware interpretation of the model's predictions.

# 3 Results and Discussion

## 3.1 Machine Learning Model Selection

Across nearly all polymers, non-linear and local learners substantially outperform linear models. Kernel SVMs, random forests, and k-nearest neighbours consistently achieve the highest coefficients of determination, while ordinary least squares and penalised linear models perform worst, aligned to previous studies [45,53,54]. For example, for CA the best models improve $R^2$ from ≈0.59 with linear regression to ≈0.97 with kernel SVM/k-NN/random forest, and for PMMA the improvement is even more pronounced (≈0.41 to ≈0.99). Nylon-6 is already well modelled by simple relations ($R^2$≈0.97) but still benefits modestly from flexible learners (≈0.984). These patterns are consistent with the strongly non-linear dependence of electrospinning outcomes on solution properties, process parameters, and environment, as well as frequent interactions among these factors as previously reported [19,52]. Ensembles outperform single trees because they reduce variance and capture higher-order interactions [55], and local methods excel when the dataset contains clusters of near-replicated conditions typical of literature compilations [56].

Figure 3 shows the "Predicted vs Observed" scatter plots for PVA obtained with linear regression and random forest for the following conditions: concentration of the solution 12%, needle diameter 20 G, rotational speed 2000 rpm, voltage 25 kV, flow rate 1 ml/h and tip-to-collector distance 11 cm. "Predicted vs Observed" scatter plots for the rest of the models and polymers for those conditions (112 cases) can be found in the links provided in Supplementary Material.

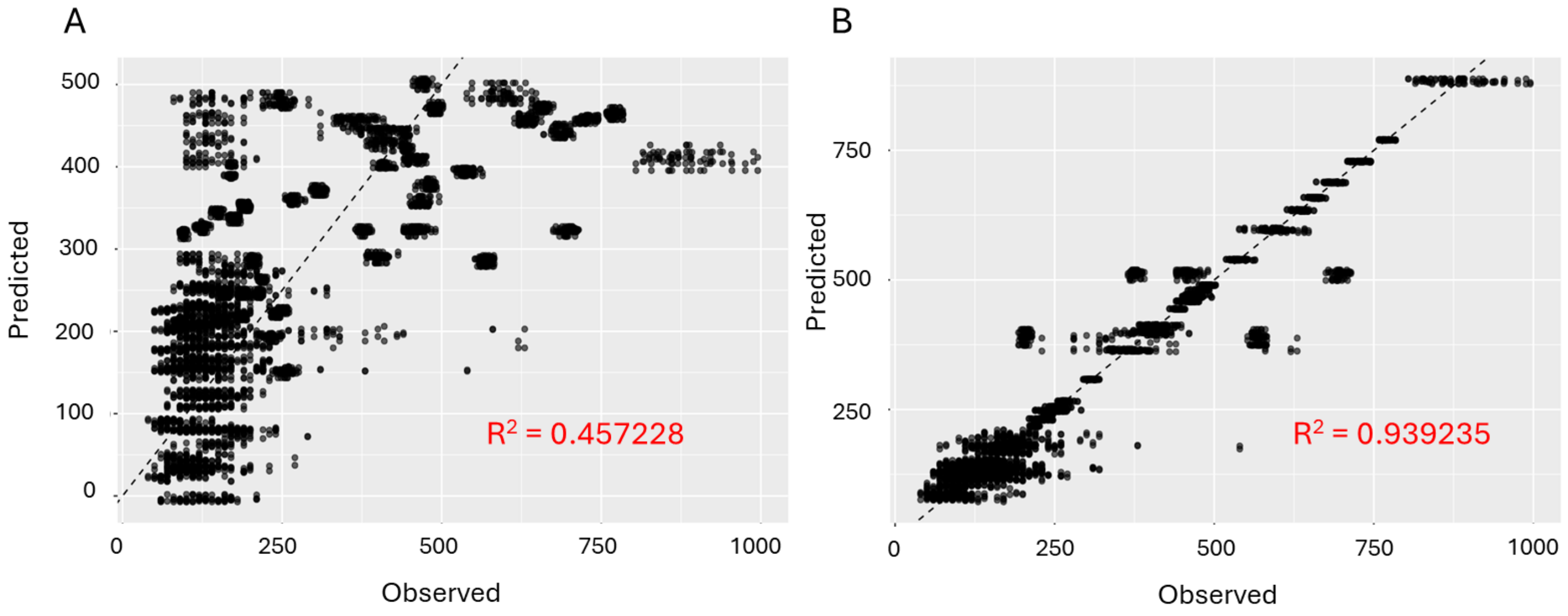

**Figure 3**. "Predicted vs Observed" scatter plots for PVA. A) Linear model (lm), B) Randon forest (ranger)

The difficulty of the prediction task is polymer-dependent. Polymers such as CA, Nylon-6, PAN, PDLLA, PMMA, PS, PU, PVA and PVDF, PU show high predictability (best $R^2$ >0.91), suggesting that the routinely reported features capture much of the governing physics for these systems. In contrast, PCL, PET, PLA and PVP intermediate ceilings (best $R^2$ ≈0.66–0.87), indicative of greater heterogeneity or missing drivers. A third group, gelatin, sulfonated PEEK and γ-PGA, remains challenging (best $R^2$<0.59). These are mostly aqueous, biopolymeric, or polyelectrolytic systems whose electrospinning is highly sensitive to variables that are inconsistently reported or absent from the feature set, including pH, ionic strength and counter-ion identity, protein conformation or gel strength (e.g., Bloom value), degree of hydrolysis, and subtle thermal or aging histories. When key physicochemical descriptors are unobserved, even flexible learners can only recover a fraction of the variance [57,58].

Two additional data characteristics explain performance variation. First, the operating range differs by polymer. Some materials are explored within tight, internally consistent windows (e.g., PET), producing nearly linear responses where all models converge to similar accuracy; others span multiple regimes (beaded vs uniform fibres; different solvent families), which reduces the effectiveness of a single global model [59]. Second, literature-derived datasets are prone to outliers and measurement heterogeneity [60]. Non-parametric methods (forests, k-NN) reduce sensitivity to these effects, which helps explain their superior performance on polymers with broad or multimodal distributions (e.g., PS and PVA) [48].

Metric choice also influences interpretation. Because the scale of the response differs markedly by polymer, RMSE and MAE are not comparable across materials; $R^2$ provides the appropriate cross-polymer lens. Within a polymer, however, RMSE/MAE remain valuable for ranking models and for practical error budgeting. Table 1 shows the evaluation of ML models for each polymer studied to predict fibre diameter for a hypothetical condition of: concentration of the solution 12%, needle diameter 20 G, rotational speed 2000 rpm, voltage 25 kV, flow rate 1 ml/h and tip-to-collector distance 11 cm.

| Polymer | Model | RMSE | MAE | $R^2$ |
|---|---|---|---|---|
| CA | lm | 175.017 ± 4.230 | 135.138 ± 3.561 | 0.588 ± 0.025 |
| | glmnet | 174.882 ± 8.602 | 135.196 ± 7.362 | 0.587 ± 0.025 |
| | rpart | 53.992 ± 4.224 | 42.526 ± 3.442 | 0.960 ± 0.006 |
| | earth | 138.544 ± 4.509 | 112.891 ± 3.421 | 0.742 ± 0.015 |
| | svmRadial | 46.031 ± 3.001 | 31.351 ± 1.726 | 0.972 ± 0.003 |
| | knn | 45.100 ± 1.675 | 32.138 ± 1.059 | 0.972 ± 0.002 |
| | ranger | 45.036 ± 1.560 | 32.125 ± 1.182 | 0.973 ± 0.003 |
| Gelatin | lm | 210.855 ± 17.442 | 158.071 ± 8.131 | 0.324 ± 0.033 |
| | glmnet | 210.919 ± 11.164 | 158.619 ± 4.039 | 0.323 ± 0.018 |
| | rpart | 174.602 ± 16.526 | 97.377 ± 4.569 | 0.538 ± 0.041 |
| | earth | 185.908 ± 14.775 | 115.994 ± 5.950 | 0.475 ± 0.034 |

| Polymer | Model | RMSE | MAE | $R^2$ |
|---|---|---|---|---|
| PLA | lm | 105.616 ± 4.475 | 90.255 ± 4.409 | 0.573 ± 0.051 |
| | glmnet | 105.605 ± 7.327 | 90.247 ± 6.583 | 0.573 ± 0.045 |
| | rpart | 93.487 ± 6.894 | 59.481 ± 6.742 | 0.656 ± 0.062 |
| | earth | 93.791 ± 7.388 | 61.310 ± 7.191 | 0.657 ± 0.065 |
| | svmRadial | 97.569 ± 8.583 | 54.123 ± 6.819 | 0.640 ± 0.082 |
| | knn | 93.556 ± 5.912 | 59.344 ± 5.799 | 0.661 ± 0.054 |
| | ranger | 93.546 ± 6.501 | 59.324 ± 7.555 | 0.662 ± 0.064 |
| PMMA | lm | 421.675 ± 10.313 | 347.970 ± 5.606 | 0.412 ± 0.033 |
| | glmnet | 421.624 ± 8.050 | 347.737 ± 5.128 | 0.411 ± 0.026 |
| | rpart | 179.724 ± 13.974 | 114.973 ± 7.638 | 0.893 ± 0.019 |
| | earth | 167.878 ± 3.199 | 107.896 ± 1.959 | 0.907 ± 0.006 |
| | svmRadial | 182.572 ± 29.801 | 81.143 ± 9.241 | 0.509 ± 0.086 |
| | knn | 170.621 ± 16.482 | 83.267 ± 6.172 | 0.558 ± 0.043 |
| | ranger | 169.780 ± 18.120 | 83.204 ± 5.575 | 0.562 ± 0.049 |
| Nylon-6 | lm | 41.416 ± 1.499 | 32.094 ± 1.197 | 0.975 ± 0.001 |
| | glmnet | 41.352 ± 1.874 | 32.245 ± 1.457 | 0.975 ± 0.002 |
| | rpart | 36.642 ± 1.626 | 27.654 ± 1.383 | 0.980 ± 0.002 |
| | earth | 39.579 ± 1.807 | 30.621 ± 1.432 | 0.977 ± 0.002 |
| | svmRadial | 34.780 ± 1.449 | 24.989 ± 0.672 | 0.982 ± 0.002 |
| | knn | 32.825 ± 1.453 | 23.652 ± 0.894 | 0.984 ± 0.002 |
| | ranger | 32.828 ± 1.088 | 23.608 ± 0.973 | 0.984 ± 0.001 |
| PAN | lm | 42.335 ± 0.774 | 33.848 ± 0.651 | 0.495 ± 0.025 |
| | glmnet | 42.357 ± 0.883 | 33.884 ± 0.546 | 0.495 ± 0.019 |
| | rpart | 27.805 ± 3.716 | 20.475 ± 2.044 | 0.778 ± 0.064 |
| | earth | 21.031 ± 0.430 | 14.947 ± 0.314 | 0.875 ± 0.008 |
| | svmRadial | 15.156 ± 1.056 | 9.592 ± 0.377 | 0.935 ± 0.009 |
| | knn | 10.135 ± 0.362 | 7.420 ± 0.273 | 0.971 ± 0.002 |
| | ranger | 10.134 ± 0.269 | 7.414 ± 0.169 | 0.971 ± 0.002 |
| PCL | lm | 92.204 ± 3.472 | 68.002 ± 1.682 | 0.642 ± 0.029 |
| | glmnet | 92.172 ± 2.730 | 67.945 ± 1.420 | 0.643 ± 0.023 |
| | rpart | 85.521 ± 4.407 | 62.085 ± 2.278 | 0.691 ± 0.037 |
| | earth | 84.845 ± 2.870 | 60.618 ± 2.252 | 0.697 ± 0.038 |
| | svmRadial | 96.327 ± 9.476 | 55.808 ± 4.569 | 0.621 ± 0.072 |
| | knn | 83.941 ± 4.124 | 58.686 ± 3.328 | 0.703 ± 0.046 |
| | ranger | 83.859 ± 4.332 | 58.655 ± 3.162 | 0.702 ± 0.032 |
| | svmRadial | 55.279 ± 2.739 | 41.527 ± 1.845 | 0.990 ± 0.001 |
| | knn | 51.712 ± 1.807 | 33.234 ± 0.817 | 0.991 ± 0.001 |
| | ranger | 51.508 ± 2.970 | 33.198 ± 1.674 | 0.991 ± 0.001 |
| PS | lm | 402.950 ± 10.395 | 280.919 ± 8.958 | 0.817 ± 0.012 |
| | glmnet | 403.784 ± 20.076 | 282.010 ± 15.397 | 0.818 ± 0.012 |
| | rpart | 315.280 ± 14.132 | 224.176 ± 12.015 | 0.887 ± 0.021 |
| | earth | 301.847 ± 10.771 | 189.908 ± 5.453 | 0.898 ± 0.009 |
| | svmRadial | 307.081 ± 18.610 | 172.441 ± 10.530 | 0.898 ± 0.019 |
| | knn | 288.616 ± 8.266 | 170.740 ± 5.874 | 0.906 ± 0.010 |
| | ranger | 287.562 ± 11.511 | 169.987 ± 8.905 | 0.906 ± 0.012 |
| PU | lm | 29.170 ± 1.611 | 23.390 ± 1.218 | 0.920 ± 0.008 |
| | glmnet | 29.209 ± 0.860 | 23.393 ± 0.879 | 0.919 ± 0.005 |
| | rpart | 22.217 ± 1.640 | 17.610 ± 1.554 | 0.953 ± 0.007 |
| | earth | 25.484 ± 1.636 | 20.254 ± 1.649 | 0.938 ± 0.008 |
| | svmRadial | 19.919 ± 1.064 | 15.398 ± 0.667 | 0.962 ± 0.004 |
| | knn | 20.077 ± 1.751 | 15.440 ± 1.233 | 0.962 ± 0.007 |
| | ranger | 19.948 ± 1.068 | 15.370 ± 0.788 | 0.962 ± 0.005 |
| PVA | lm | 145.549 ± 3.264 | 111.057 ± 3.462 | 0.457 ± 0.014 |
| | glmnet | 145.486 ± 4.232 | 110.924 ± 2.948 | 0.457 ± 0.023 |
| | rpart | 102.509 ± 7.247 | 70.095 ± 4.239 | 0.728 ± 0.042 |
| | earth | 96.495 ± 3.475 | 68.832 ± 2.607 | 0.761 ± 0.013 |
| | svmRadial | 69.530 ± 6.672 | 33.611 ± 2.281 | 0.879 ± 0.020 |
| | knn | 48.814 ± 3.802 | 26.061 ± 1.767 | 0.939 ± 0.011 |
| | ranger | 48.645 ± 3.518 | 25.990 ± 1.634 | 0.939 ± 0.008 |
| PDLLA | lm | 500.809 ± 29.659 | 424.237 ± 22.662 | 0.599 ± 0.037 |
| | glmnet | 501.425 ± 19.616 | 424.735 ± 19.056 | 0.597 ± 0.030 |
| | rpart | 202.659 ± 26.015 | 147.514 ± 23.967 | 0.932 ± 0.021 |
| | earth | 467.554 ± 10.667 | 382.354 ± 11.061 | 0.649 ± 0.027 |
| | svmRadial | 57.879 ± 1.058 | 55.374 ± 1.217 | 0.996 ± 0.001 |
| | knn | 12.559 ± 0.413 | 9.365 ± 0.394 | 0.9997 ± 0.000 |
| | ranger | 12.466 ± 0.913 | 9.342 ± 0.651 | 0.9998 ± 0.000 |
| PEEK-sulfonated | lm | 20.431 ± 0.564 | 16.474 ± 0.547 | 0.130 ± 0.037 |
| | glmnet | 20.432 ± 0.868 | 16.474 ± 0.544 | 0.130 ± 0.042 |
| | rpart | 17.030 ± 1.029 | 12.966 ± 0.970 | 0.396 ± 0.068 |
| | earth | 15.838 ± 0.631 | 12.363 ± 0.442 | 0.481 ± 0.018 |
| | svmRadial | 14.399 ± 1.169 | 10.497 ± 0.698 | 0.573 ± 0.051 |
| | knn | 13.975 ± 0.780 | 10.459 ± 0.533 | 0.594 ± 0.037 |
| | ranger | 14.044 ± 0.762 | 10.517 ± 0.503 | 0.589 ± 0.037 |
| PET | lm | 63.312 ± 5.541 | 46.441 ± 4.977 | 0.869 ± 0.018 |
| | glmnet | 63.311 ± 4.166 | 46.441 ± 3.825 | 0.867 ± 0.011 |
| | rpart | 63.324 ± 2.605 | 46.449 ± 2.230 | 0.868 ± 0.008 |
| | earth | 63.379 ± 4.766 | 46.506 ± 3.928 | 0.869 ± 0.013 |
| | svmRadial | 64.561 ± 2.598 | 46.641 ± 1.864 | 0.868 ± 0.008 |
| | knn | 63.274 ± 3.234 | 46.483 ± 2.472 | 0.869 ± 0.007 |
| | ranger | 63.265 ± 2.033 | 46.472 ± 1.450 | 0.869 ± 0.007 |
| PVDF | lm | 178.836 ± 3.638 | 127.426 ± 2.748 | 0.577 ± 0.018 |
| | glmnet | 178.731 ± 3.880 | 127.601 ± 1.800 | 0.578 ± 0.026 |
| | rpart | 135.819 ± 8.902 | 95.328 ± 6.621 | 0.756 ± 0.037 |
| | earth | 85.685 ± 3.224 | 55.733 ± 3.164 | 0.903 ± 0.007 |
| | svmRadial | 74.342 ± 2.267 | 41.827 ± 1.383 | 0.927 ± 0.004 |
| | knn | 73.161 ± 1.753 | 40.885 ± 0.952 | 0.929 ± 0.004 |
| | ranger | 73.161 ± 2.683 | 40.907 ± 1.596 | 0.929 ± 0.005 |
| PVP | lm | 1670.019 ± 45.876 | 1093.773 ± 25.227 | 0.349 ± 0.014 |
| | glmnet | 1670.128 ± 58.132 | 1090.075 ± 24.867 | 0.349 ± 0.008 |
| | rpart | 1112.460 ± 59.527 | 572.430 ± 32.495 | 0.712 ± 0.043 |
| | earth | 1522.597 ± 78.302 | 964.072 ± 28.101 | 0.459 ± 0.012 |
| | svmRadial | 1238.841 ± 105.595 | 562.267 ± 19.778 | 0.674 ± 0.038 |
| | knn | 1029.798 ± 63.047 | 492.726 ± 26.714 | 0.755 ± 0.021 |
| | ranger | 1024.187 ± 68.943 | 491.455 ± 30.790 | 0.757 ± 0.031 |
| γ-PGA | lm | 88.995 ± 7.094 | 53.577 ± 3.928 | 0.171 ± 0.033 |
| | glmnet | 89.040 ± 9.351 | 53.641 ± 4.561 | 0.169 ± 0.059 |
| | rpart | 86.653 ± 4.156 | 51.524 ± 3.042 | 0.214 ± 0.024 |
| | earth | 85.276 ± 7.114 | 48.853 ± 4.064 | 0.236 ± 0.035 |
| | svmRadial | 90.236 ± 7.340 | 43.736 ± 4.610 | 0.192 ± 0.046 |
| | knn | 85.060 ± 8.853 | 47.764 ± 5.122 | 0.242 ± 0.045 |
| | ranger | 85.042 ± 9.607 | 47.702 ± 4.558 | 0.241 ± 0.048 |

**Table 1.** Evaluation of ML models for each polymer studied to predict fibre diameter distributions (mean ± standard deviation).

### 3.2. Importance of the features on the distribution of the diameter of the fibres

Each model for each polymer provides different importance of the features. As an illustrative example, this section presents the importance plot, SHAP value distribution, and correlation matrix computed for the Support Vector Machine (SVM) model for PVDF under the following conditions: concentration of the solution 12%, needle diameter 20 G, rotational speed 2000 rpm, voltage 25 kV, flow rate 1 ml/h and tip-to-collector distance 11 cm. The importance of each feature on the diameter of the fibres was studied for the seven ML models and the 16 polymers for the same electrospinning settings (112 cases), and they can be found in Supplementary Material.

Figure 4 shows the variable-importance plot, computed for the support vector machine (SVM) model for PVDF. This plot shows a clear hierarchy at the global level. Solution concentration is by far the most influential predictor of fibre diameter. Needle diameter comes next, followed by rotation speed, with smaller contributions from flow rate and voltage. The tip-to-collector distance and the type of collector contributes very little in this ranking. For an SVM, this importance reflects how much model accuracy drops when each predictor is perturbed. It therefore captures overall influence on predictions but not directionality. The ordering is physically sensible: concentration governs viscosity and chain entanglement, needle diameter affects the emerging jet at the nozzle, and rotation speed can draw fibres after formation, whereas flow rate, voltage, and distance usually have more context-dependent effects.

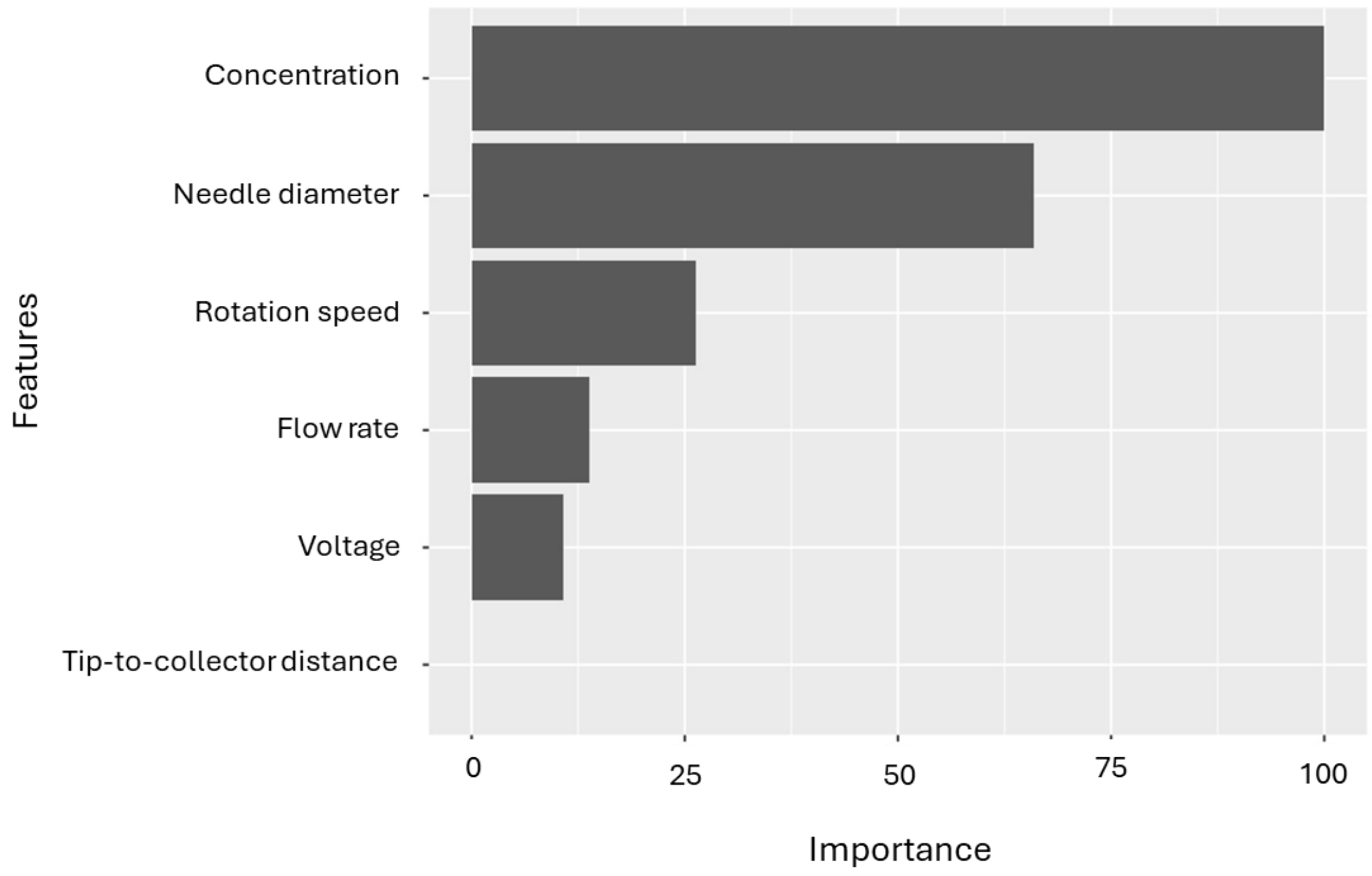

**Figure 4**. Feature importance in electrospun fibre diameter for PVDF and SVM model.

The SHAP summary complements this result by describing how each variable moves individual predictions up or down. Figure 5 shows that solution concentration shows the broadest spread of SHAP values, confirming that changes in concentration consistently drive large shifts in the predicted fibre diameter across the dataset. Flow rate and tip-to-collector distance display wider SHAP dispersions than their global ranks might suggest, indicating that they matter in specific regions of the operating space even if their average effect is smaller. Voltage has a moderate, fairly uniform impact. Rotation speed and needle diameter are clustered near zero for most observations, with occasional large contributions. This pattern indicates that these two variables tend to influence the SVM decision function only under certain combinations of the other settings. The bidirectional spreads around zero further point to non-linear responses and interactions, which the SVM (especially with a non-linear kernel) captures naturally.

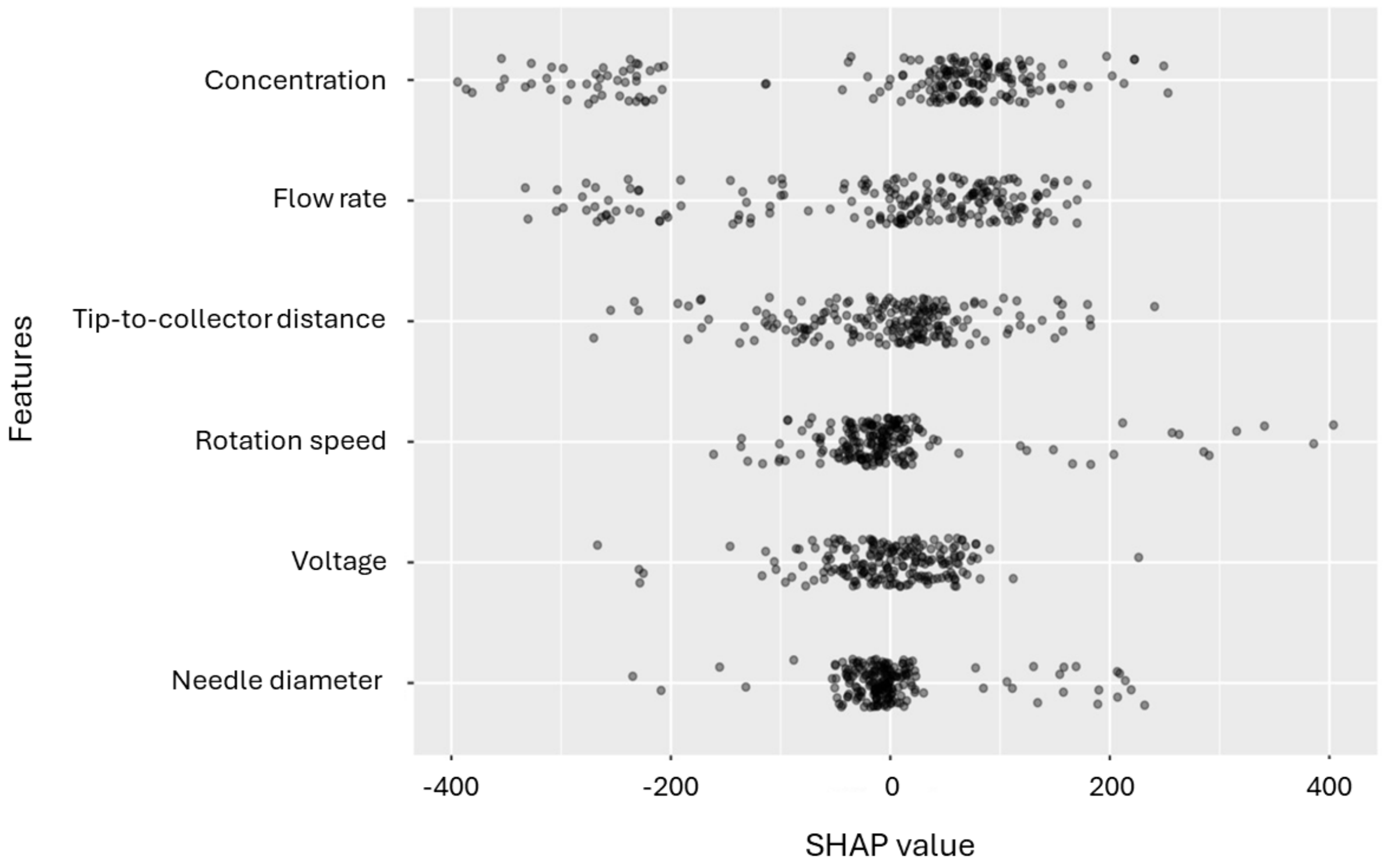

**Figure 5**. SHAP values distribution for each feature for PVDF and SVM model.

The correlation matrix summarizes linear associations (Figure 6). With fibre diameter, the strongest Pearson correlations are for solution concentration (≈0.48) and needle diameter (≈0.39), followed by rotation speed (≈0.25), voltage (≈0.16), and flow rate (≈0.13). Tip-to-collector distance shows a weak negative correlation (≈–0.17). The matrix also reveals notable inter-predictor correlation, such as rotation speed with voltage (≈0.74), flow rate with rotation speed (≈0.69), and tip-to-collector distance with voltage (≈0.66). Such collinearity means several predictors carry overlapping information, which can dilute a variable's apparent importance or obscure its unique effect when viewed through simple pairwise correlations.

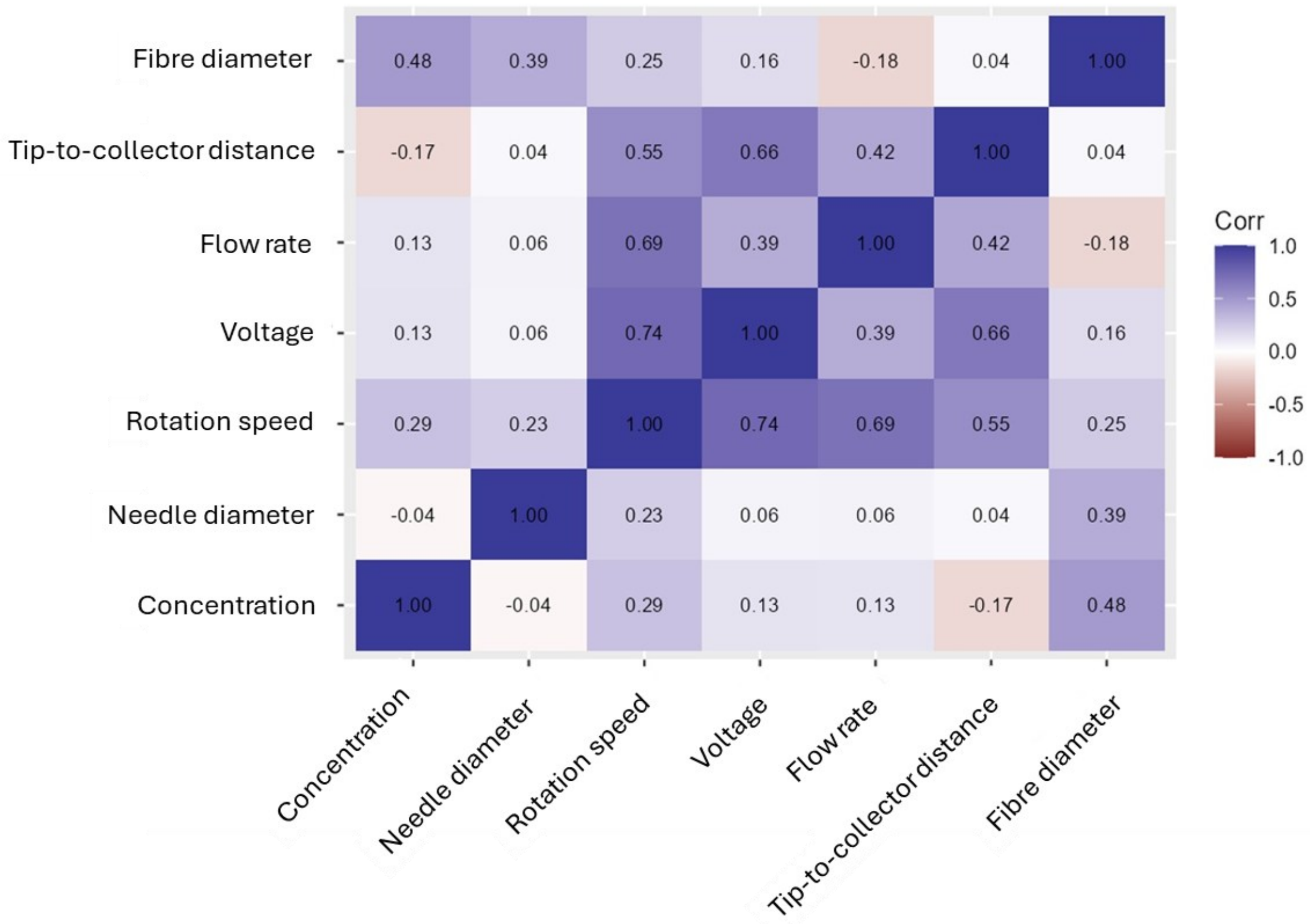

**Figure 6**. Correlation matrix of electrospinning variables for PVDF and SVM model.

Taken together, the three analyses converge on a consistent conclusion: solution concentration is the primary and most reliable parameter governing fibre diameter, in agreement with previous studies [19,52]. The dominant influence of concentration across models aligns with classical electrohydrodynamic theory, wherein concentration dictates chain entanglement and viscoelastic stress, key factors that suppress bead formation and enhance jet stability. Needle diameter also emerges as an influential parameter, ranking highly in global SVM importance, although its SHAP values are near zero in most cases, suggesting that its effect is pronounced only within specific regions of the design space. This observation is consistent with its known influence on the initial jet radius and shear stress at the nozzle, which in turn affect the onset of bending instabilities. In contrast, the tip-to-collector distance exhibits a weak overall importance and modest linear correlation but displays distinct local SHAP effects, indicating that its impact arises under particular combinations of voltage and flow rate. Voltage and rotation speed show secondary yet context-dependent effects, with the latter influencing post-jet drawing and extensional strain during flight. These findings collectively support the strong machine learning, identified importance of solution concentration and needle diameter, as well as the conditional influence of rotation speed, voltage, and flow rate.

To further understand how each processing parameter interacts with solution concentration, the 3D response-surface plots reveal clear and physically coherent trends that align with the SVM importance ranking, SHAP distributions, and correlation analysis (Figure 7).

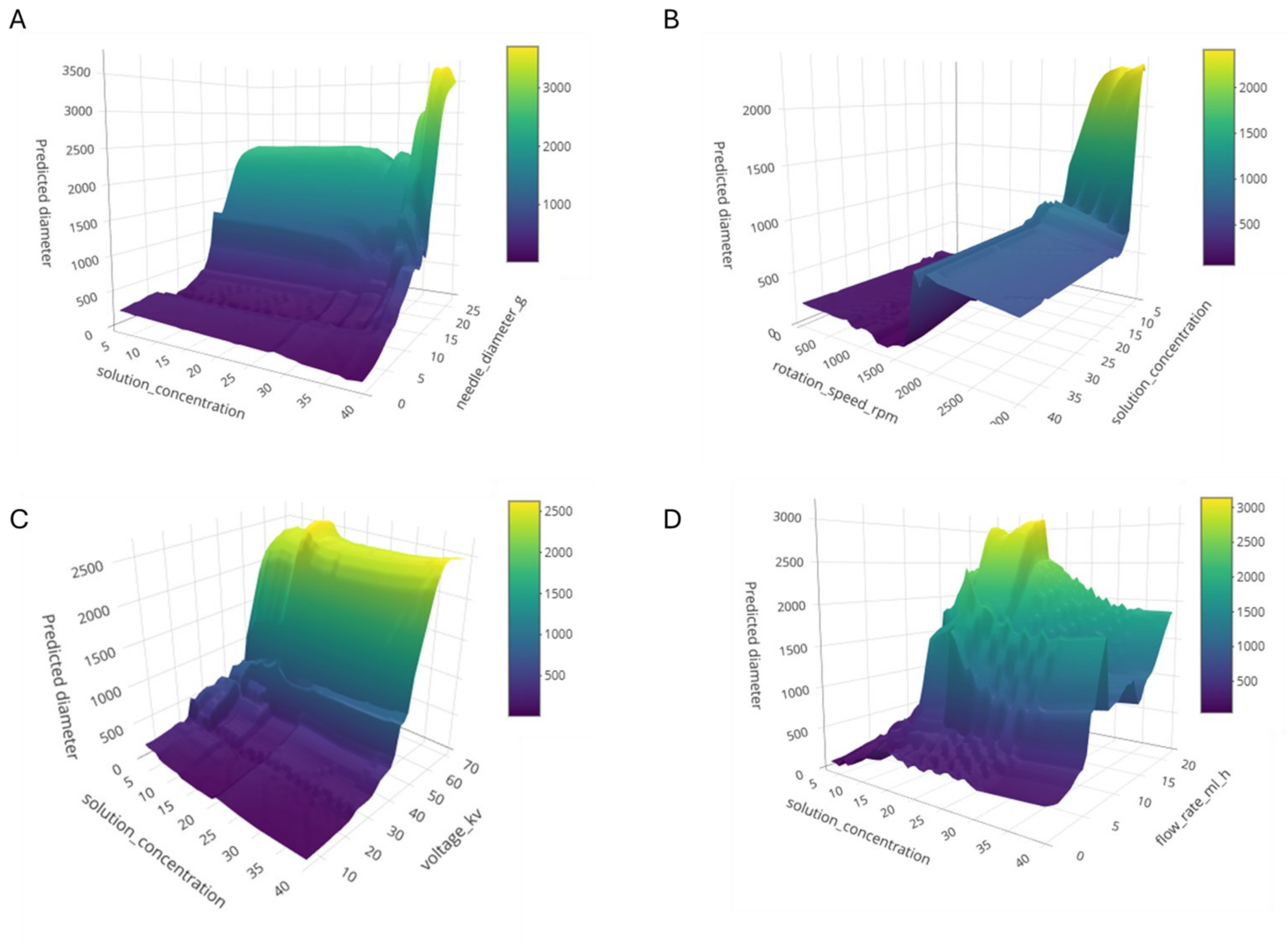

**Figure 7**. 3D response-surface plots with predicted fibre diameter (nm) as a function of solution concentration and A) needle diameter, B) rotation speed, C) voltage, and D) flow rate.

The concentration–needle diameter surface shows the sharpest increases in fibre diameter, with needle effects becoming significant only at moderate-to-high concentrations, supporting its high SVM importance and its locally activated SHAP contributions. The concentration–rotation speed surface remains mostly flat at low concentrations but shows modest thinning at higher concentrations, consistent with the parameter's moderate global importance and its context-dependent SHAP variability. The concentration–voltage surface exhibits smooth, uniform gradients, reflecting voltage's moderate and steady influence predicted by the SHAP summary and its relatively weak linear correlation. In contrast, the concentration–flow rate surface displays irregular, localised diameter increases at high concentrations, explaining why flow rate ranks low in global importance yet shows wider SHAP dispersion in specific operating regions. Together, these surfaces visually confirm that concentration is the dominant and most robust predictor of fibre diameter, while needle diameter, rotation speed, voltage, and flow rate exert secondary but regime-specific effects, precisely the hierarchy and interaction patterns indicated by the ML and correlation analyses.

From a practical perspective, solution concentration should be tightly controlled, while the remaining parameters should be adjusted with consideration of their interactions.

### 3.3. FibreCastML Open Web Platform

To broaden access and encourage collaboration in the electrospinning community, a polymer-specific Shiny application (FibreCastML) was developed to predict nanofibre diameters from experimentally controllable process conditions. The interface is intentionally simple. Users first select the polymer and the collector type, then enter the operating parameters: solution concentration, needle diameter (g), rotation speed, voltage (kV), flow rate (ml $h^{-1}$), and tip-to-collector distance (cm). On running a prediction, the app returns not only a point estimate but also the cross-validated distribution of predicted diameters, providing a practical sense of variability. The tool further supports decision-making by highlighting inputs that fall outside the empirical ranges observed for the chosen polymer, proposing solvent combinations (and median ratios when available) drawn from similar historical experiments, and presenting transparent diagnostics, including predicted-versus-observed scatter, metrics ($R^2$, RMSE, MAE), global variable importance, SHAP-based local explanations, and a correlation heat map. The web interface of FibreCastML is shown in Figure 8. The web FibreCastML can be accessed at https://electrospinning.shinyapps.io/electrospinning/

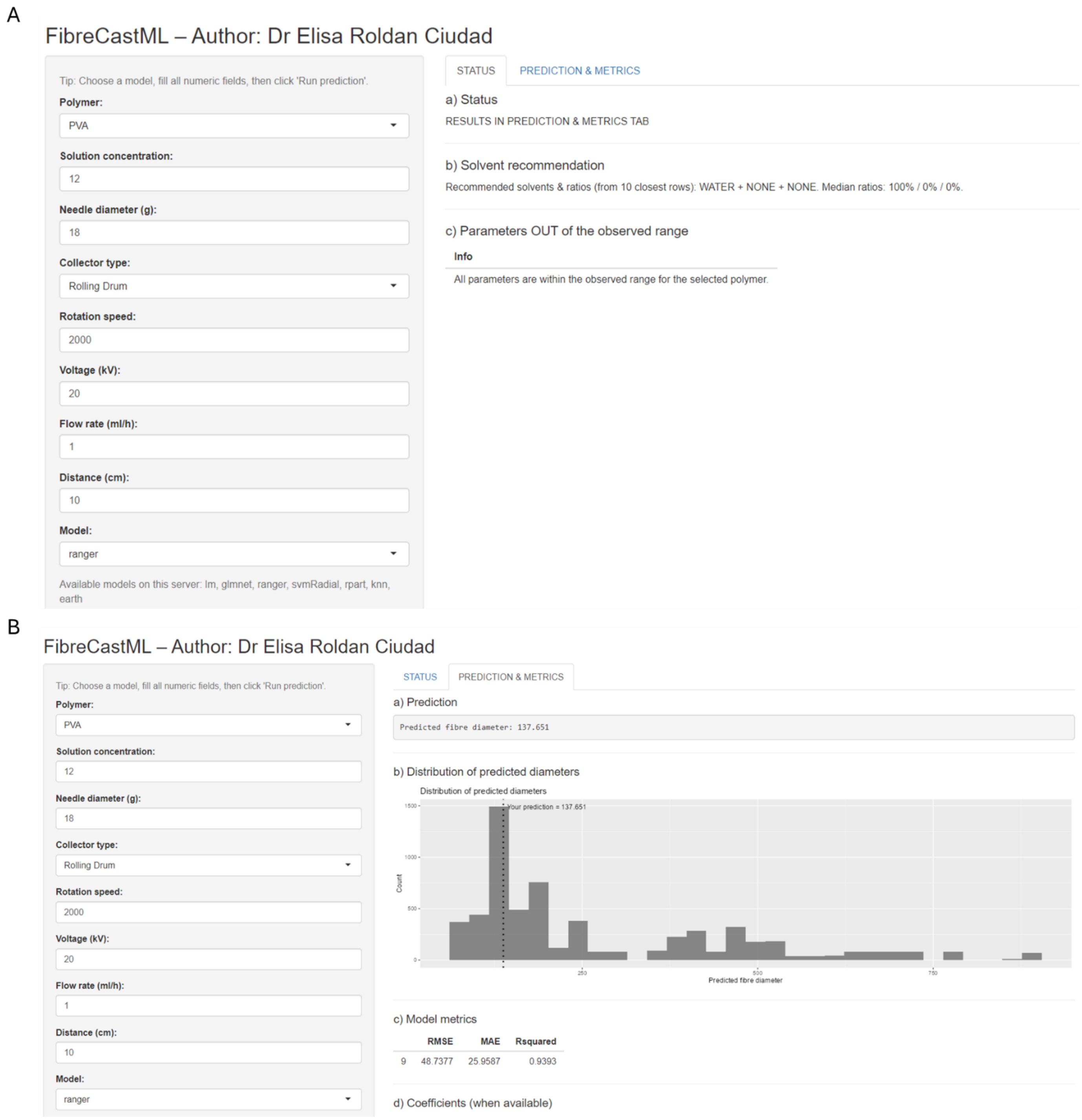

**Figure 8.** Screenshots of the web FibreCastML interface. A) Status tab, B) Prediction and metrics tab

All preprocessing and model training are performed within the app, and a complete Excel report, capturing inputs, predictions, metrics, interpretability outputs, and publication-ready figures, can be downloaded for audit and sharing (Excel reports can be found in Supplementary Material). In typical use, if an exact setting is unknown, users may supply a random set up and the out-of-range panel and distribution plots then contextualise the plausibility of those choices. By consolidating data, analytics, and interpretability into a single, user-friendly environment, the application lowers the barrier to applying advanced modelling in electrospinning labs, improves comparability across studies, and accelerates experimental planning while encouraging reproducible, transparent practice.

### 3.4. Experimental case study

Statistical analyses were conducted to quantitatively evaluate the agreement between FibreCastML-predicted fibre diameters and the experimentally measured diameters obtained from the TL-01 and Spraybase® electrospinning systems. Prior to selecting appropriate tests, the distribution of fibre diameters for each group (Predicted, TL-01, Spraybase®) was assessed for normality using the Shapiro–Wilk test. All groups demonstrated significant deviations from a normal distribution (TL-01: $p = 0.0015$; Spraybase®: $p = 0.00235$; Predicted: $p = 3.36\times10^{-9}$), confirming the need for non-parametric statistical methods.

Descriptive statistics were first computed to characterise the central tendency and dispersion of each dataset. TL-01 fibres exhibited a median diameter of 148.5 ± 30.24 nm, Spraybase® fibres showed a slightly lower median of 137.75 ± 28.76 nm, and FibreCastML prediction obtained with the same process parameters was 137.651 nm. The overall ranges were comparable between electrospinning systems and the single fibre diameter prediction.

Figure 9 shows the location of the predicted fibre diameter within the distribution of predictions from cross-validation obtained from FibreCastML. Figure 10 shows examples of SEM images and total fibre diameter distributions of the scaffolds created with TL-01 and Spraybase®.

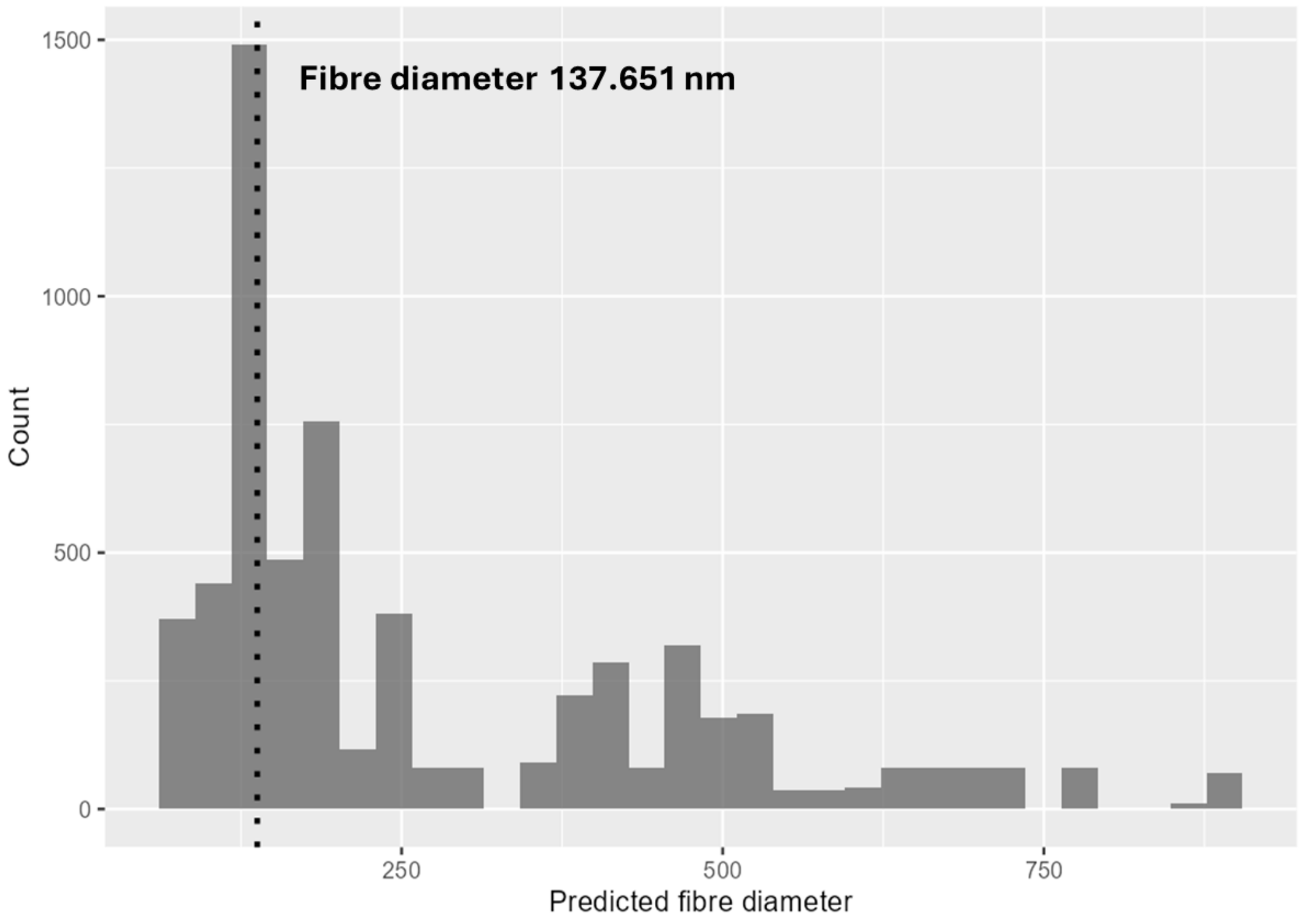

**Figure 9.** Predicted fibre diameter obtained from the web FibreCastML interface

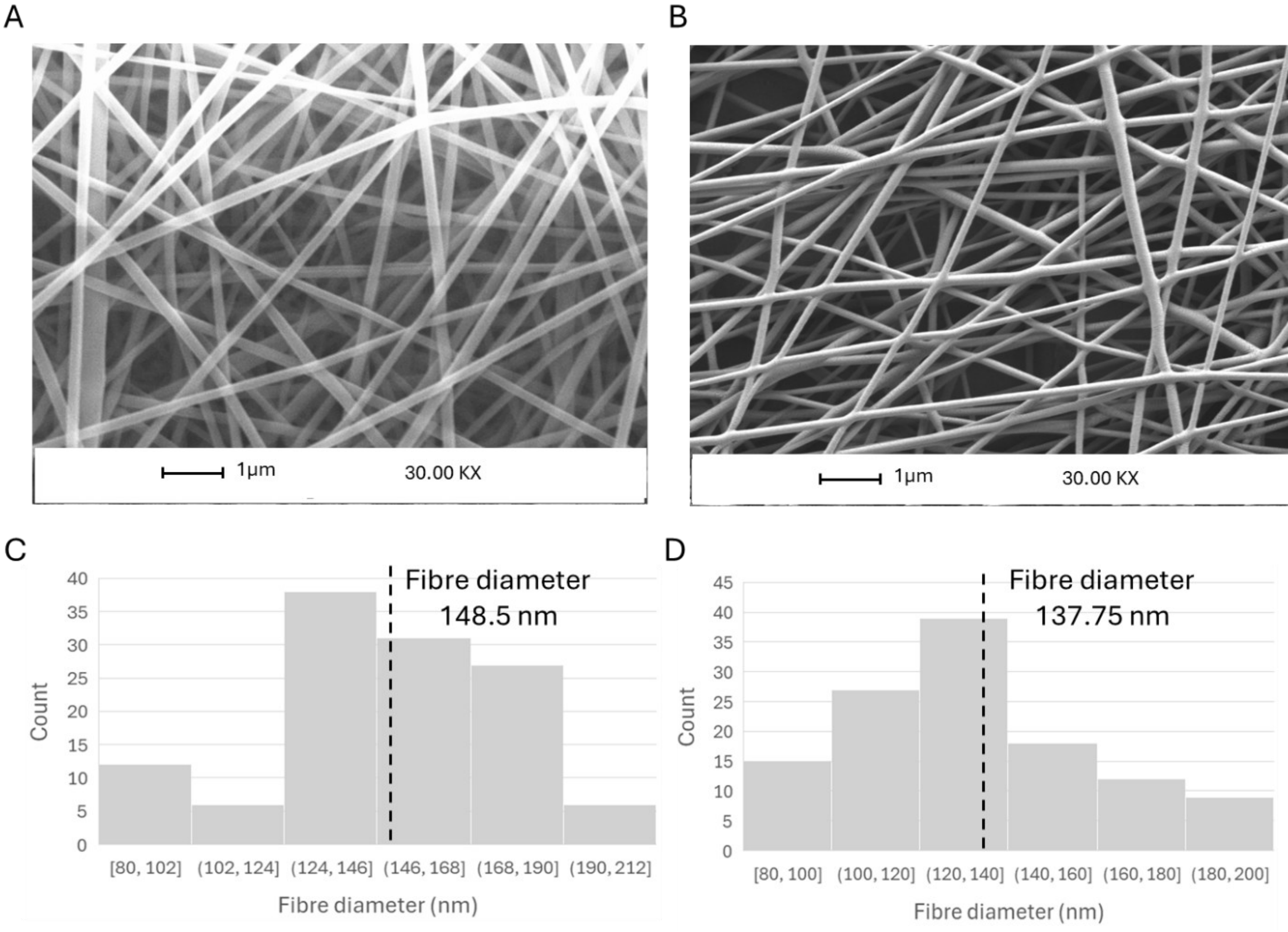

**Figure 10.** SEM images and total fibre diameter distributions created with A and C) TL-01, B and D) Spraybase®

The comparison between the fibre diameter distribution obtained with the scaffolds produced with Spraybase® electrospinner and the predicted diameter distribution using the residual bootstrap reveals substantial agreement in overall shape, central tendency, and general dispersion of both populations. From an inferential perspective, the three statistical tests applied (Kolmogorov–Smirnov (KS), Mann–Whitney U, and independent-samples t-test) consistently failed to detect significant differences between the real and simulated distributions (KS: $p = 0.13$; UMW: $p = 0.32$; $t$-test: $p = 0.34$). These results indicate that the two samples are statistically compatible in terms of distribution, median, and mean. Complementary metrics reinforce this conclusion: the Kullback–Leibler divergence (KL) was 0.1497, the Wasserstein distance was 15.73, and the overlap coefficient (OVL) reached 84.11%, all pointing to a high degree of similarity between the two distributions. Overlaid density plots show that both distributions share a common modal region around 120–140 µm, where the highest probability density largely coincides (Figure 11). The model accurately reproduces the core structure of the process, capturing the prominent concentration of diameters in this range. Although the simulated distribution exhibits slightly heavier tails (particularly towards larger diameters) most of the probability mass remains within the experimentally observed bounds.

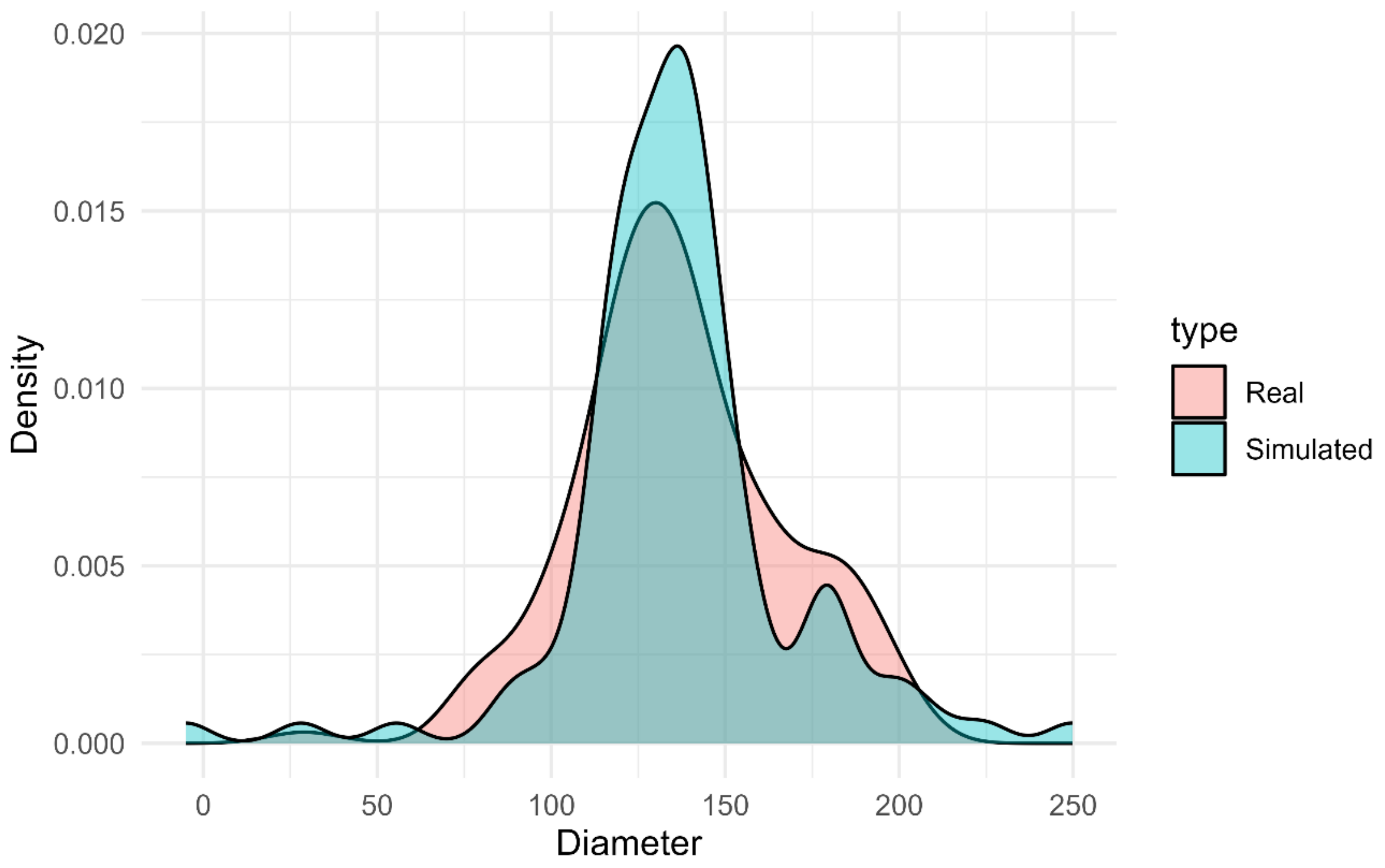

**Figure 11.** Fibre diameter distributions from real scaffolds created with Spraybase® (Real) and predicted distribution (Simulated).

Finally, violin plots confirm that the median, interquartile range, and skewness of the simulated distribution closely approximate those of the real distribution (Figure 12). The additional variability observed at the extremes of the simulated violin is consistent with the deliberate incorporation of real-world variability through the Residual Bootstrapping procedure, designed to preserve the inherent heterogeneity of electrospinning and avoid an artificially overoptimistic fit.

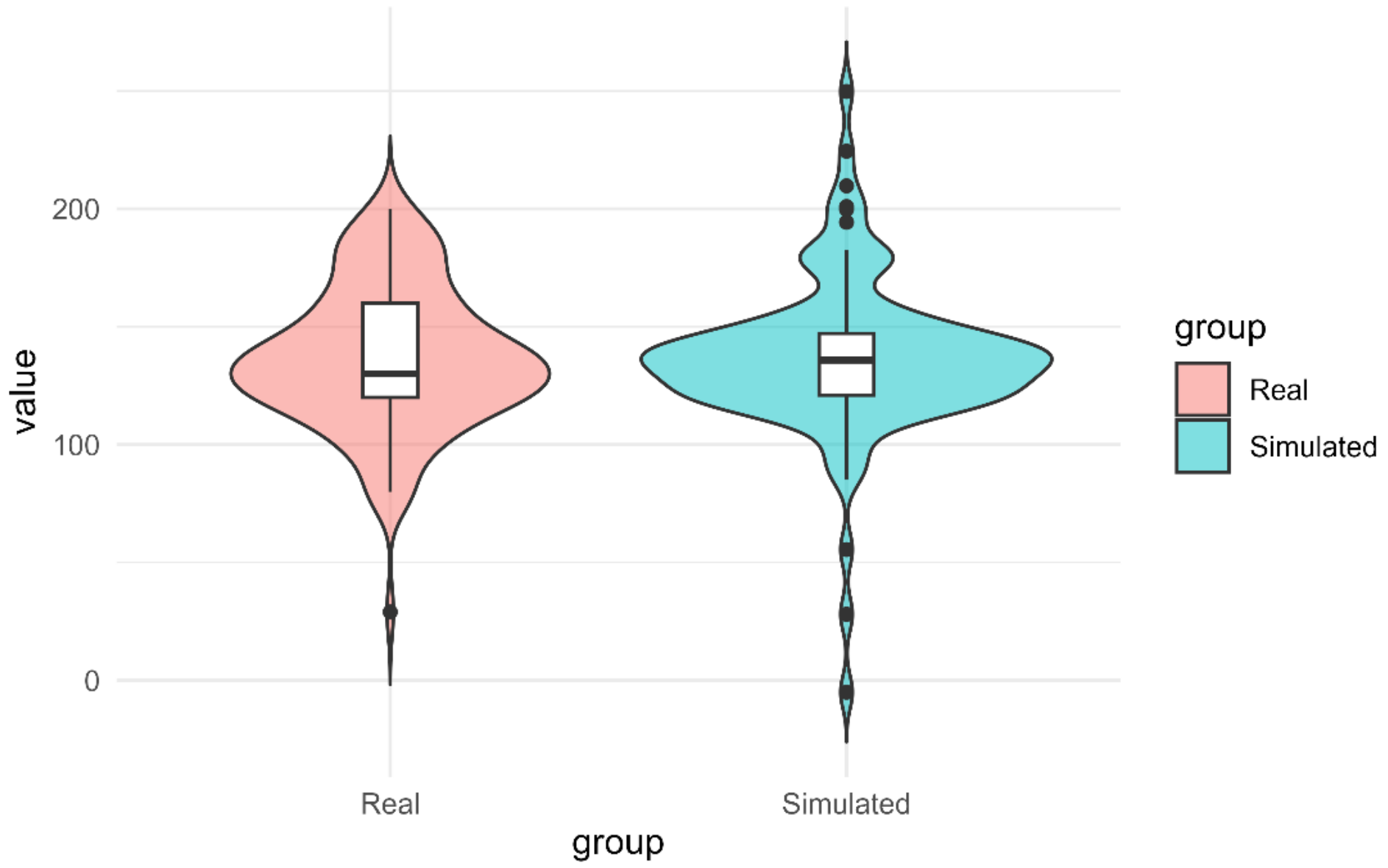

**Figure 12.** Violin comparison between real values obtained with Spraybase® (Real) and predicted values (Simulated).

The comparison between the fibre distribution obtained with the scaffolds produced with TL-01 and the predicted revealed a moderate-to-high level of global concordance between the real and simulated fibrillar diameter distributions, as indicated by an overlap coefficient (OVL) of 73.97%. This suggests that approximately three-quarters of the simulated distribution coincides with the experimental one. Shape and distance metrics further support this interpretation: the Kullback–Leibler divergence (KL) was 0.261, reflecting a low divergence, while the Wasserstein distance was 16.67, indicating a moderate separation between distributions. However, hypothesis tests consistently rejected equality between distributions: Kolmogorov–Smirnov ($p = 1.4\times10^{-5}$), Mann–Whitney U ($p = 0.00127$), and t-test ($p = 0.0224$), indicating significant differences in shape, median, and mean. Visual analyses corroborate these findings: density plots reveal that the simulated distribution is narrower with a sharper peak, while the real distribution exhibits heavier tails; and violin plots show greater spread in the real data and a slight bias towards smaller diameters in the simulated set (Figure 13). Overall, FiberCastML captures the central tendency and approximately 74% of the real distribution's structure, yet notable discrepancies persist in form and dispersion, differences attributable to uncontrolled physical factors in electrospinning (e.g., humidity, temperature, microflow variations, jet instability) that the model cannot fully reconstruct. Nevertheless, the system provides realistic simulations for diameter prediction, preserving the essential statistical features of the process.

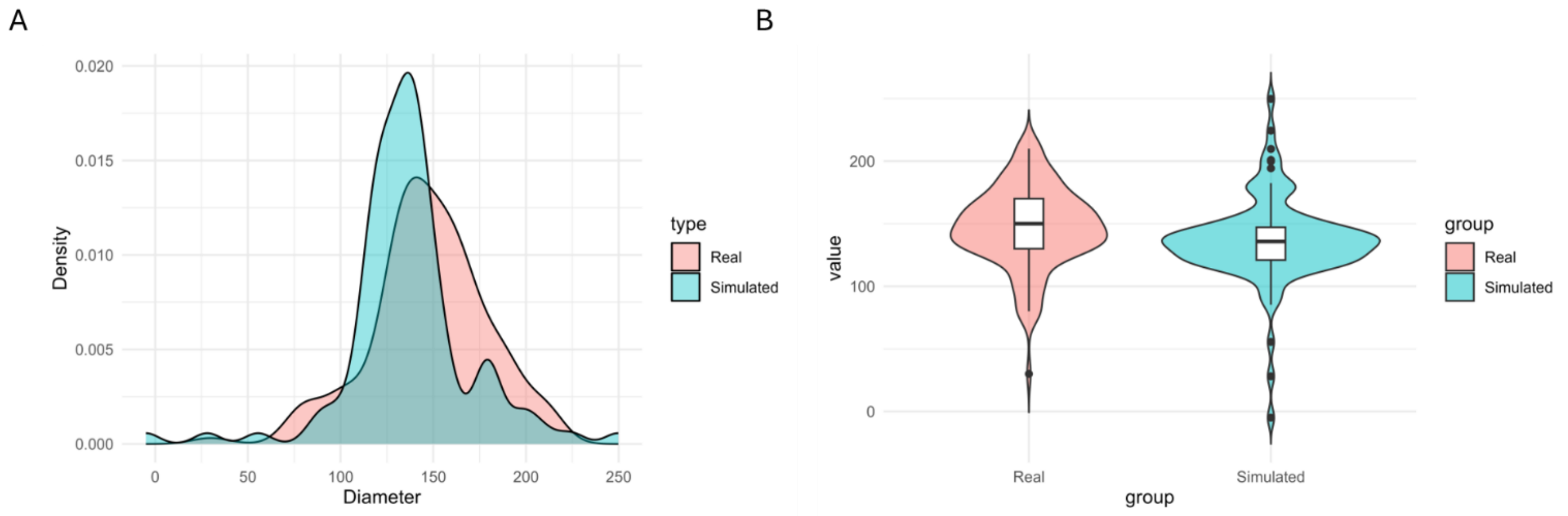

**Figure 13.** Density and violin plot for real values obtained with LT-01 (Real) and predicted values (Simulated).

In summary, external case study using two independent datasets (Spraybase® and TL-01, $N = 124$ each) demonstrates that FiberCastML effectively reproduces the statistical structure of electrospun PVA fibre diameters under diverse experimental conditions. The model consistently captures central tendencies and key distributional features, providing strong predictive validity despite the inherent stochasticity of the electrospinning process.

### 3.5. Limitations of the study

Recent studies reported that tree- or rule-based algorithms such as GBM and Cubist deliver the lowest prediction errors and the highest coefficients of determination, owing to their capacity to model non-linear relationships, exploit ensemble strategies, and maintain robustness against overfitting [44,45]. In our study, these models achieved marginally higher $R^2$ values and lower error metrics than alternative approaches; however, they could not be deployed reliably in the server-side Shiny environment due to version and platform inconsistencies between development and production (R and package binaries). For transparency and reproducibility, they were therefore neither included in the code nor were their results reported.

Occasionally, it is necessary to reload the page to establish a connection to the Shiny back end. This behaviour reflects conditions in the hosting provider's infrastructure that lie outside our control.

The experimental validation study has several limitations that should be acknowledged. First, although the external validation dataset encompassed multiple electrospinning systems and imaging modalities, all experiments and image acquisitions were performed by a single researcher. As a result, potential researcher-to- researcher variability in sample preparation and imaging was not assessed. Secondly, the dataset, while deliberately diverse, remains relatively limited in size compared with the range of conditions that may be encountered across different laboratories. Additional validation with a larger, multicentric dataset involving independent users would further strengthen the evidence for the FibreCastML's generalisability. Finally, only PVA-based scaffolds were evaluated in this study; therefore, the model's performance on other polymer systems remains to be established.

Only routinely reported and consistently available variables were retained for analysis, including solution concentration, needle diameter, rotation speed, voltage, flow rate, and tip-to-collector distance. Parameters such as viscosity, electrical conductivity, surface tension, molecular weight, humidity, material's grade and brands, deposition time and material of the collector were excluded due to their limited reporting (currently appearing in fewer than 12% of studies included in our database) which hindered meaningful cross-study generalisation and introduced substantial data missingness. However, incorporating physicochemical and rheological parameters can provide valuable insight into the governing electrohydrodynamic mechanisms [22]. Therefore, future work will aim to include these predictors, along with polymer combinations, to enhance model robustness and improve the understanding of factors influencing fibre formation.

# 4 Conclusions

This study advances electrospinning optimisation from mean-focused to distribution-aware prediction, establishing a world-first open framework (FibreCastML) capable of forecasting full fibre-diameter spectra from standard experimental parameters. Trained on a uniquely curated database of 68,538 measurements across 16 polymers, the framework combines rigorous nested cross-validation and interpretability analysis to deliver reliable, transparent, and generalisable insights.

Across polymers, non-linear and local learners consistently outperform linear baselines, confirming the importance of nonlinearity in electrospinning behaviour. Solution concentration remains the dominant factor, providing a robust lever for design control. In the experimental case study, FibreCastML accurately reproduced the measured PVA diameter distribution (Kolmogorov–Smirnov $p > 0.13$ and overlap coefficient of 84%), allowing to cut significantly experimental iterations and solvent consumption, demonstrating tangible laboratory and environmental impact.

Operationally, the app transforms complex ML workflows into accessible tools for practitioners by offering: (i) Polymer-specific distribution forecasts and diagnostics ($R^2$, RMSE, MAE); (ii) Interpretability outputs (variable importance, SHAP) linking parameters to outcomes; (iii) Automated error checking and solvent recommendations from historical conditions; and (iv) Downloadable, auditable reports to support reproducibility and cross-lab comparability.
These features position FibreCastML as a decision engine for data-driven, sustainable electrospinning. Future versions will integrate curated physicochemical and rheological features to further enhance generalisability and extend to polymer blends and multi-objective optimisation.

In summary, distribution-aware machine learning transforms electrospinning into a predictive, sustainable, and globally shareable process, reducing waste and accelerating the design of biomimetic scaffolds for tissue engineering, wound dressings, and drug delivery. This framework redefines experimental strategy in electrospinning, marking a paradigm shift toward reproducible and environmentally responsible nanomanufacturing.

## Data availability statement

The data supporting this article is available in Supplementary Material.

## Author Contributions

Conceptualization, literature review, coding and app development, data curation, data analysis, statistical and machine learning studies and writing the article: ER

Initial concept and review the article: KA, SMR, GC, NDR

Review the article: RF, RE, TS

## Funding

This research received no external funding.

## Additional Information

**Competing Interests.** The authors declare that the research was conducted in the absence of any commercial or financial relationships that could be construed as a potential conflict of interest.

**Supplementary Material.** The online version contains available supplementary material to support the results and discussion section. The web FibreCastML can be accessed at https://electrospinning.shinyapps.io/electrospinning/